\documentclass[sigconf, nonacm]{acmart}

\usepackage{microtype}
\usepackage{graphicx}
\usepackage{subfigure}
\usepackage{booktabs} %

\usepackage{hyperref}
\usepackage{setspace}

\usepackage{amsmath}
\usepackage{mathtools}
\usepackage{amsthm}
\usepackage{nicefrac}       %
\usepackage{xcolor}         %

\usepackage{tikz}
\usetikzlibrary{graphs,graphs.standard}
\usetikzlibrary{positioning,chains,fit,shapes,calc}

\usepackage{enumitem}
\usepackage{stmaryrd}

\usepackage{multirow} %
\usepackage{tabularx}
\usepackage{multicol}

\usepackage[capitalize,noabbrev]{cleveref}
\usepackage{url}            %
\usepackage{booktabs}       %

\usepackage{algorithmic}
\usepackage{algorithm}

\theoremstyle{plain}
\newtheorem{theorem}{Theorem}[section]

\newtheorem{lemma}[theorem]{Lemma}

\theoremstyle{definition}
\newtheorem{definition}[theorem]{Definition}

\theoremstyle{remark}

\usepackage[textsize=tiny]{todonotes}

\usepackage{pdfpages}
\usepackage{makecell}
\usepackage{subfiles}
\usepackage{caption}
\usepackage{csvsimple}
\usepackage{arydshln}
\usepackage[round-mode=places, round-integer-to-decimal, round-precision=4,
    table-format = 1.4, 
    table-number-alignment=center,
    round-integer-to-decimal]{siunitx}
\usepackage{booktabs} %
\usepackage{microtype}

\usepackage{colortbl}

\newcommand\BigO{\mathcal{O}}

\usepackage[symbol]{footmisc}
\renewcommand{\thefootnote}{\fnsymbol{footnote}}

\usepackage{booktabs}

\usepackage{pgfplots}
\pgfplotsset{compat=1.5}

\newcommand{\R}{\mathbb{R}}

\newcolumntype{N}{c@{}S}

\newcommand{\Powerset}[1]{\mathcal{P}(#1)}

\author{Lukas Gianinazzi$^{1,2}$* , Maximilian Fries$^1$*, Nikoli Dryden$^1$, \\ Tal Ben-Nun$^1$, Maciej Besta$^1$, Torsten Hoefler$^1$}
\affiliation{\department{$^1$Department of Computer Science} \institution{ETH Zurich}}

\begin{document}

\fancyhead{}

\title{Learning Combinatorial Node Labeling Algorithms}

\begin{abstract}

We present a novel neural architecture to solve graph optimization problems where the solution consists of arbitrary node labels, allowing us to solve hard problems like graph coloring.
We train our model using reinforcement learning, specifically policy gradients, which gives us both a greedy and a probabilistic policy.
Our architecture builds on a graph attention network and uses several inductive biases to improve solution quality.
Our learned deterministic heuristics for graph coloring give better solutions than classical degree-based greedy heuristics and only take seconds to apply to graphs with tens of thousands of vertices. 
Moreover, our probabilistic policies outperform all greedy state-of-the-art coloring baselines and a machine learning baseline. 
Finally, we show that our approach also generalizes to other problems by evaluating it on minimum vertex cover and outperforming two greedy heuristics.

\end{abstract}

\maketitle

\section{Introduction}

\footnotetext[1]{Equal Contribution.}
\renewcommand*{\thefootnote}{\arabic{footnote}}
\footnotetext[2]{Corresponding Author: lukas.gianinazzi@inf.ethz.ch}

Combinatorial optimization problems on graphs, such as graph coloring and minimum
vertex cover, are the subject of numerous works
in academia and industry.
Graph coloring (GC) has many real-world applications, including scheduling
problems~\cite{Marx03graphcolouring, DBLP:series/sci/Myszkowski08}, register allocation~\cite{Chaitinregalloc, DBLP:conf/pldi/SmithRH04},
and mobile network autoconfiguration~\cite{DBLP:conf/iwcmc/BandhCS09}.
Minimum vertex cover (MVC) is a well-studied problem in algorithmic graph
theory~\cite{DBLP:conf/soda/OnakRRR12, DBLP:conf/spaa/GhaffariJN20, DBLP:conf/soda/BhattacharyaHN17} with applications including text
summarization~\cite{Islam2017ApplicationOM} and computational biology~\cite{mvc_biology}.
For these problems, the common goal is to \emph{assign labels to nodes subject to
combinatorial feasibility constraints and costs}. Therefore, we can generalize these problems into a single problem we call \emph{combinatorial node labeling}, which includes many 
other problems such as traveling salesman~\cite{DBLP:journals/ior/DantzigFJ54,GareyNP-Complete}, maximum cut~\cite{Kar72}, and list coloring~\cite{jensen1995graph}.

As combinatorial node labeling is NP-hard~\cite{Kar72,GareyNP-Complete}, many machine learning approaches
have been proposed to solve special cases.
A learning scheme faces two fundamental challenges when compared to classical heuristics:
(1) it should match or outperform heuristics in quality of solution and performance; and
(2) it should generalize across graphs of different sizes.
Recent work addressed these challenges for the traveling
salesman problem~\cite{bello2017neural, cappart2020combining, drori2020learning, ma2019combinatorial}, influence
maximization~\cite{DBLP:conf/nips/ManchandaMDMRS20}, and minimum vertex cover~\cite{17Dai, li18gcntreesearch}.
For graph coloring, \citet{lemos} introduce an estimator for the chromatic number $\chi(G)$ of a graph, but do not construct a coloring and can over- or underestimate $\chi(G)$ (i.e., there may not be a feasible solution with the estimated number of colors).

\begin{figure}
    \centering
    \includegraphics[width=1.0\linewidth]{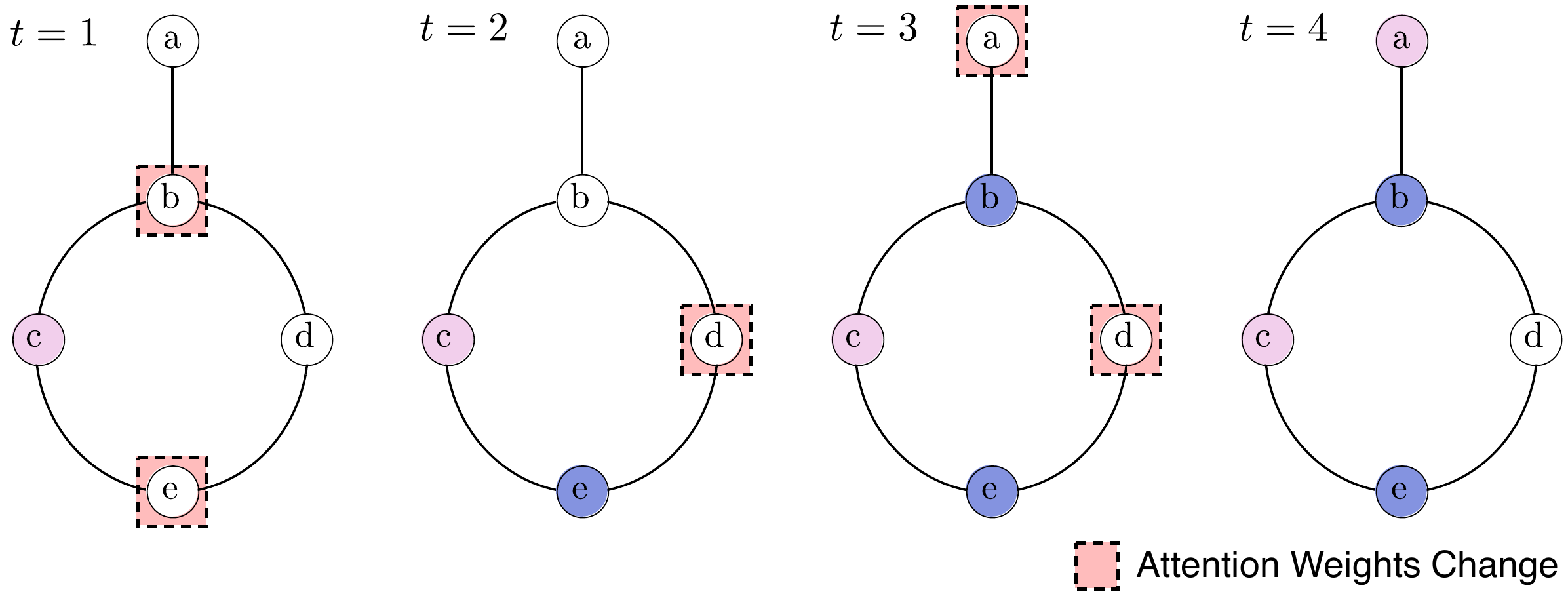}
     \caption{\emph{Spatial locality of the decoding.} After labeling a node, only its neighbors' attention weights change. The example shows how a graph is $2$-colored using the vertex order $c,e,b,a,d$. The nodes whose attention weights change have a box around them. For example, when the first node $c$ is colored, only its neighbors $b$ and $e$ receive new attention weights. The figure omits the last step (where $d$ is colored).} %
    \label{fig:Decoder}
\end{figure}

Existing machine learning methods do not easily generalize to cases where the number of labels is not known in advance, such as graph coloring.
To address this limitation, we use policy gradient reinforcement learning~\cite{sutton,kool2019attention} to learn a node ordering and combine this with a simple label rule to label each node according to the ordering. We show that for the chosen label rules, there still exists an order that guarantees an optimal solution. 
Our model uses a graph neural network (GNN) encoder~\cite{DBLP:conf/iclr/VelickovicCCRLB18} and an attention-based~\cite{lee2018attention} decoder to assign weights for which node to label next. These weights can either be used greedily or interpreted as probabilities. Our decoder incorporates a \emph{temporal locality} inductive bias, where the selection of a node is conditioned only on the previously labeled node and a global graph context.
Further, we also introduce a \emph{spatial locality} inductive bias, whereby labeling a node only impacts the weights of its neighbors. See \Cref{fig:Decoder} for an illustration.

We present the combinatorial node labeling framework and evaluate it on the graph coloring and minimum vertex cut problems. We introduce a generic GNN architecture that demonstrates significantly improved results for neural graph coloring and outperforms two state-of-the-art greedy heuristics for minimum vertex cover. A qualitative analysis of the learned heuristics reveals their capability to adapt depending on the properties of the test graph. Our ablation studies show that the introduced temporal and spatial biases improve test scores.

\subsection{Related Work}

\textbf{Supervised learning}\; Recent approaches like \citet{joshi} and \citet{DBLP:conf/nips/ManchandaMDMRS20} obtain good results for influence maximization (IM) and the traveling salesman problem (TSP), respectively. Both approaches use supervised learning. Supervised learning is more sample efficient than reinforcement learning and can lead to overall better results. The fundamental downside is that it is not applicable to every problem. First, it can be difficult to formulate a problem in a supervised manner, since it might have many optimal solutions (e.g., GC). Second, even if the problem admits a direct supervised formulation, we still need labeled data for training, which can be hard to generate and relies on an existing solver. For IM, the approach of~\citet{DBLP:conf/nips/ManchandaMDMRS20} shows promising results on graphs much larger than those seen in training. For TSP, the approach of~\citet{joshi} is very efficient but does not generalize well to graphs larger than those seen in training. \citet{li18gcntreesearch} also use supervised learning and produce good results on minimum vertex cover (MVC), maximum independent set, and maximal clique. 

\textbf{Reinforcement learning}\; \citet{17Dai} provide a general framework for learning problems like MVC and TSP that is trained with reinforcement learning. It shows good results across different graph sizes for the covered problems, but is not fast enough to replace existing approaches.
\citet{kool2019attention} focus on routing problems like TSP and the vehicle routing problem. They outperform \citeauthor{17Dai} on TSP instances of the training size. Unfortunately, their approach does not seem to generalize to graph sizes that are very different from those used for training.
Several other reinforcement learning approaches have been proposed and evaluated for TSP~\cite{bello2017neural, cappart2020combining, drori2020learning, ma2019combinatorial}. \citet{barrett2019exploratory} consider the maximum cut (MaxCut) problem.~\citet{DBLP:journals/corr/abs-1902-10162} present a Monte Carlo search tree approach specialized only for graph coloring.

These methods do not address the general node labeling framework, but instead model the solution as a permutation of vertices (e.g., TSP, vehicle routing) or a set of nodes or edges (e.g., MVC, MaxCut).

\section{Combinatorial Node Labeling}

We introduce \emph{combinatorial node labeling}, which generalizes many important hard graph optimization problems, including graph coloring (see \Cref{apx:node-labeling} for a list).

We consider an undirected, unweighted, and simple graph $G=(V, E)$ with $n$ nodes in $V$ and $m$ edges in $E$. We denote the neighbors of a node $v$ by $N(v)$. We assume w.l.o.g. that the graph is connected and hence $m=\Omega(n)$.  %

A node labeling is a mapping $ c:V \to \{0, \dotsc, n\} $. A \emph{partial node labeling} is a mapping $c' : V' \to \{0, \dotsc, n\}$ for any subset of nodes $V'\subseteq V$. A node labeling problem is subject to a feasibility condition and a real-valued \emph{cost function} $f$. The cost function maps a node labeling $c$ to a real-valued cost $f(c)$.
We require that the feasibility condition is expressed in terms of an efficient (polynomial-time computable) \emph{extensibility test} $T:\Powerset{V \times \{0, \dotsc, n\}} \times V \times \{0, \dotsc, n\} \to \{0,1\}$, where $\mathcal{P}$ denotes the powerset. We say the extensibility test passes when it returns $1$.

Intuitively, given a partial node labeling $c':V' \to \{0, \dotsc, n\}$, a node $v\not \in V'$, and label $\ell$, the extensibility test passes if and only if the partial node labeling $c'$ can be extended by labeling node $v$ with $\ell$ such that the partial node labeling can be extended into a node labeling.
Formally, the extensibility test characterizes the set of \emph{feasible solutions}: 
\begin{definition}\label{def:feasibility}
A node labeling $c$ is feasible if and only if there exists a sequence of node-label pairs $(v_1, \ell_1), \dotsc, (v_n, \ell_n)$ such that for all $i\geq 0$ the extensibility test $T$ satisfies $$T(\{(v_1, \ell_1), \dotsc, (v_i, \ell_i)\},  v_{i+1}, \ell_{i+1})=1 \enspace .$$
\end{definition}

The goal of the node labeling problem is to \emph{minimize} the value of the cost function among the feasible node labelings.
For consistency, an infeasible node labeling has infinite cost.

Next, we present the two node labeling problems on which we focus in our evaluation.

\subsection{Graph coloring (GC)}

\begin{definition}
     A $k$-coloring of a graph \(G = (V,E) \) is a node labeling \(c:V \to \{1, 2, \dots, k\} \) such that no two neighbors have the same label, i.e.,  \( \forall \{u,v\} \in E : c(u)~\neq~c(v).\)
\end{definition}
The cost function for GC is the number of distinct labels (or colors) $k$. Given a partial node labeling $c' : V' \to \{1, \dotsc, k \}$ and any vertex-label pair $(v, \ell)$, the extensibility test passes for $(c', v, \ell)$ if and only if the extended partial node labeling $c'\cup(v, \ell)$ is a $k$- or $(k+1)$-coloring of the induced subgraph $G[V' \cup \{v\}]$. In particular, the test does not pass when $\ell> k+1$. The smallest $k$ for which there is a $k$-coloring of $G$ is the \emph{chromatic number} \( \chi(G)\) of $G$.

\subsection{Minimum vertex cover (MVC)}

\begin{definition}
A vertex cover of a graph \(G = (V,E) \) is a node labeling $c : V \to \{0, 1\}$ such that every edge is incident to at least one node with label $1$, i.e., $ \forall \{u,v\} \in E : c(u)=1 \lor c(v) = 1. $
\end{definition}
The cost function for MVC is the number of nodes with label $1$. Given a partial node labeling $c': V' \to \{0, 1\}$ the extensibility test passes for $(c', v, \ell)$  if and only if the extended partial node labeling $c'\cup(v, \ell)$ is a vertex cover of the induced subgraph $G[V'~\cup~\{v\}]$.

\section{Node labeling policies}\label{sec:mdp}

Next, we show how every node labeling problem can be formulated as a (finite) Markov decision process (MDP), during which nodes are successively added to a partial node labeling until a termination criterion is met. Then, we discuss how to learn a parameterized policy for this problem. In \Cref{sec:graph-learning}, we will present a GNN architecture for parameterizing a policy for such MDPs.

\subsection{Node labeling as a Markov Decision Process}

 We embed the cost function $f$ and the extensibility test into the MDP. Note that we do not require a way to measure the cost of partial node labelings. 
We formulate the state space, action space, transition function, and reward:\\
\textbf{State space}\; A state $S$ represents a partial node labeling. It is a set of pairs $S=V' \times \mathcal{L}$ for a subset of nodes $V'\subseteq V$ and a subset of labels $\mathcal{L} \subseteq \{0, \dotsc n\}$. A state is terminal if $V'=V$. Hence, the set of states is the powerset $\Powerset{V \times \{0, \dotsc, n\}}$ of the Cartesian product of the vertices and labels.\\
\textbf{Action space}\; In state $S$, the set of legal actions are the pairs $(v, \ell)$ for nodes $v$ and labels $\ell$ which pass the extensibility test of the problem for the partial node labeling given by $S$ (i.e., $T(S, v, l)=1$).\\ %
\textbf{Transition function}\; In our case, the transition function $\mathcal{T}$ is deterministic. That is, given the current state $S_t$ and an action $(v,\ell)$, $\mathcal{T}(S_t, (v, \ell))$ yields the next state $S_{t+1}= S_t \cup \{ (v,\ell) \}$.\\
\textbf{Reward}\; For a terminal state $S$ representing the node labelling $c$, the reward is $- f(c)$. For all other states, the reward is $0$. %

Note that since our tasks are episodic, the return equals the sum of the rewards (specifically the reward received in the terminal state). In particular, we do not use discounting.

In \Cref{apx:proofs:mdp}, we prove that the terminal states of this MDP correspond to the feasible solutions:
\begin{lemma}\label{lem:mdp}
For any node labeling problem, there is an MDP whose terminal states correspond to the feasible solutions with a cost equal to the negative return.
\end{lemma}

In the vast majority of reinforcement learning approaches to solve combinatorial graph optimization problems~\cite{kool2019attention, 17Dai, ma2019combinatorial, drori2020learning}, a state $S$ corresponds to a set or sequence of nodes that are already added to a solution set. Instead, in our setting the state represents a partial node labeling. This means that in addition to problems like MVC and TSP, we can also model problems with more than two labels (even when the number of labels is not known in advance). Graph coloring is such a problem. %

A \emph{policy} is a mapping from states to probabilities for each action. Note that we can turn a probabilistic policy into a deterministic \emph{greedy} policy by choosing the action with largest probability. Next, we present how to train this policy end-to-end using policy gradients.

\subsection{Policy optimization}

We train a parameterized node labeling model by policy gradients, specifically \textsc{Reinforce}~\cite{bello2017neural} with a greedy rollout baseline \cite{kool2019attention}. 
At a high level, the algorithm works as follows. 
We begin by initializing two models, the \emph{current} model and the \emph{baseline} model. For each graph in the batch, the algorithm performs a probabilistic rollout of the policy. The baseline model performs a greedy rollout. The difference between the two costs determines the policy gradient update. 
After every epoch, we perform a (one-sided) paired $t$-test over the cost on a challenge dataset to check if the baseline model should be replaced with the current model. See \Cref{apx:training-policy} for more details.

\section{Graph learning architecture}\label{sec:graph-learning}

Next, we present a graph learning approach to parameterizing policies for node labeling in our MDP framework.
Our MDP formulation is modeled after greedy node labeling algorithms. A greedy node labeling algorithm assigns a label in $\{0, \dotsc, n\}$ to one node after another based on a problem-specific heuristic. Hence, it can be seen as providing (1) an order on the nodes and (2) a rule to label the next selected node.

We focus on learning an order on the nodes and pick a label that passes the extensibility test according to a fixed rule. The following two lemmas show there exists a simple \emph{label rule} that ensures the optimal solution can be found for GC and MVC (see \Cref{apx:proofs:label} for the proofs):
\begin{lemma}\label{lem:color-order}
For every graph $G$, there exists an ordering of vertices for which choosing the smallest color that passes the extensibility test colors $G$ optimally.
\end{lemma}

\begin{lemma}\label{lem:mvc-order}
For every graph $G$, there exists an ordering of vertices for which choosing the label $1$ until every vertex in $G$ is adjacent to a node with label $1$ produces a minimum vertex cover of $G$.
\end{lemma}
We expect similar results can be obtained for most other node labeling problems.

A common strategy in many successful heuristics is to choose the order of the nodes based on their neighborhoods: The ListRight heuristic for MVC \cite{listright} assigns a node to the vertex cover based on the assignment of its neighbors. For GC, the DSATUR strategy selects nodes according to their saturation degree~\cite{DSATUR}. If a new node is selected, only the saturation degree of its neighborhood can change; the others remain unchanged.

Instead of a handcrafted ordering heuristic, we learn to assign weights to each node and choose the nodes according to their weights. To compute these weights, we introduce a novel \emph{spatial locality} inductive bias inspired by the greedy heuristics: labeling a node should only affect the weights of its neighbors. As we will show in~\Cref{sec:experiments:ablation}, this leads to better test scores compared to the alternatives of updating all or none of the weights when a node is labeled.

\subsection{Architecture overview} \label{sec:model_architecture}
Our architecture consists of an encoder and a decoder to learn a policy specific the node labeling problem. The encoder learns the local structural information that is important for the problem in the form of a node embedding. It is possible to instantiate any GNN in the encoder.

The \emph{context embedding} encapsulates information about the graph itself (enabling the network to adapt its actions to the graph), the last node that was labeled, and its label. Using only the last node and its label results in a \emph{temporal locality} inductive bias. Adding more nodes into the context embedding provided no benefit (see~\Cref{sec:experiments:ablation}).

\begin{figure}
    \centering
    \includegraphics[width=1.0\linewidth]{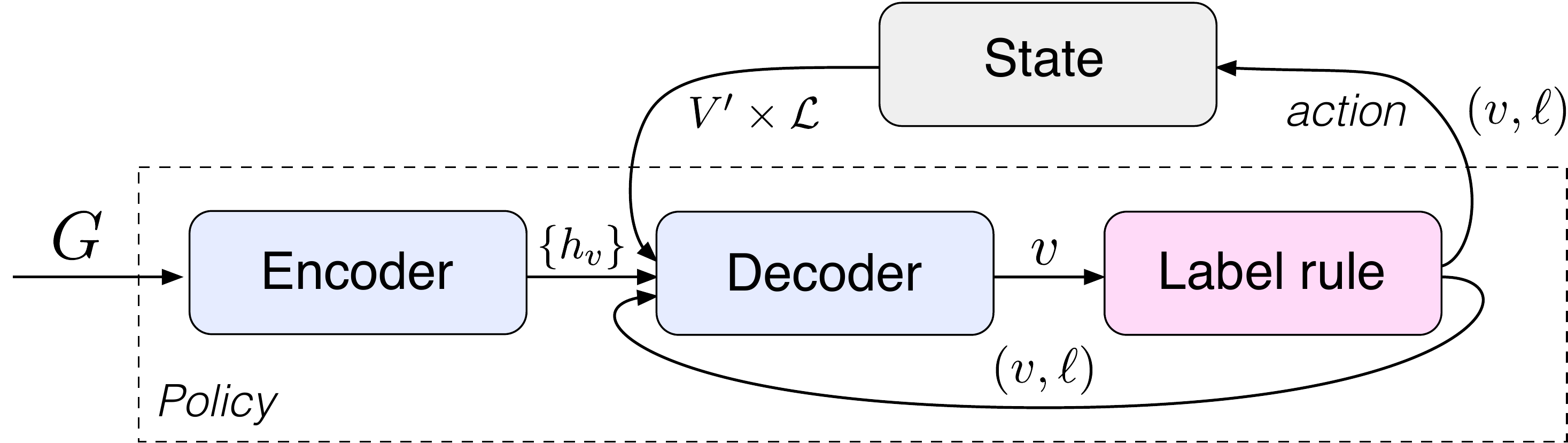}
    \caption{High-level architecture. The encoder (a GNN) reads the input graph and produces embeddings for each node. The decoder takes these embeddings, together with the current state $V' \times \mathcal{L}$ and the last taken action, and outputs the next node. The node $v$ feeds into the label rule, which produces the next action $(v, \ell)$, leading to the next state $(V' \times \mathcal{L}) \cup \{(v, \ell)\}$. This process repeats until reaching a terminal state (when all nodes are labelled). } %
    \label{fig:architecture}
\end{figure}

The decoder uses the node embeddings and the context embedding to select the next node based on attention weights between the node embeddings and the context embedding. After the decoder picks the next node $v$,  the label rule (see \Cref{lem:color-order} and \Cref{lem:mvc-order}) assigns the label $\ell$ for the node. The policy then takes the action $(v, \ell)$. Then, the context embedding is updated and the decoder is invoked again until all nodes are labelled. 
\Cref{fig:architecture} overviews our architecture.

\subsection{Node features}
Each node $v$ is associated with an input feature vector $x_v$. Our input features consist of a combination of sine and cosine functions of the node degree, similarly to positional embeddings~\cite{vaswani2017attention}. This representation ensures that input features are bounded in magnitude even for larger graphs. We subtract the mean node degree from the degrees on the synthetic dense graph instances.

\subsection{Encoder}

We use a hidden dimension of size $d$ (unless stated otherwise, $d = 64$).
The input features are first linearly transformed and then fed into a GNN, which produces, for each node $v$, a node embedding $h_v \in \mathbb{R}^{d}$.
We use a three-layer Graph Attention Network (GAT)~\cite{DBLP:conf/iclr/VelickovicCCRLB18, vaswani2017attention,lee2018attention}, additive multi-head attention with four heads, batch normalization~\cite{batch_normalization} with a skip connection \cite{skip_connections} at each encoder layer, and leaky ReLU activations~\cite{maas2013rectifier}. %

\subsection{Context embedding}

Next we describe how to construct the \emph{context embedding}, which is an additional input to the decoder. The context embedding is a function of the output of the encoder and the partial node labeling. Each label $\ell$ has a \emph{label embedding} $h_\ell$, which is a max-pooling over the embeddings of the nodes with the same label $\ell$. The graph has a \emph{graph embedding} $h_G$, which is a max-pooling over all node embeddings. %
In the context embedding, we introduce a \emph{temporal bias} by only considering the last labeled node (and the graph embedding): Specifically, we denote the node that is labeled in time step $t$ by $v^{(t)}$ and its label by $\ell^{(t)}$. Then, the \emph{context embedding} $g_t$ concatenates the following components: (1) The graph embedding $h_G$,  (2) the embedding $h_{v^{(t-1)}}$ of the last labeled node $v^{(t-1)}$, and (3) the label embedding $h_{\ell^{(t-1)}}$ of the last labeled node's label $\ell^{(t-1)}$. 

\begin{figure}[t]
    \centering
    \includegraphics[width=.99\linewidth]{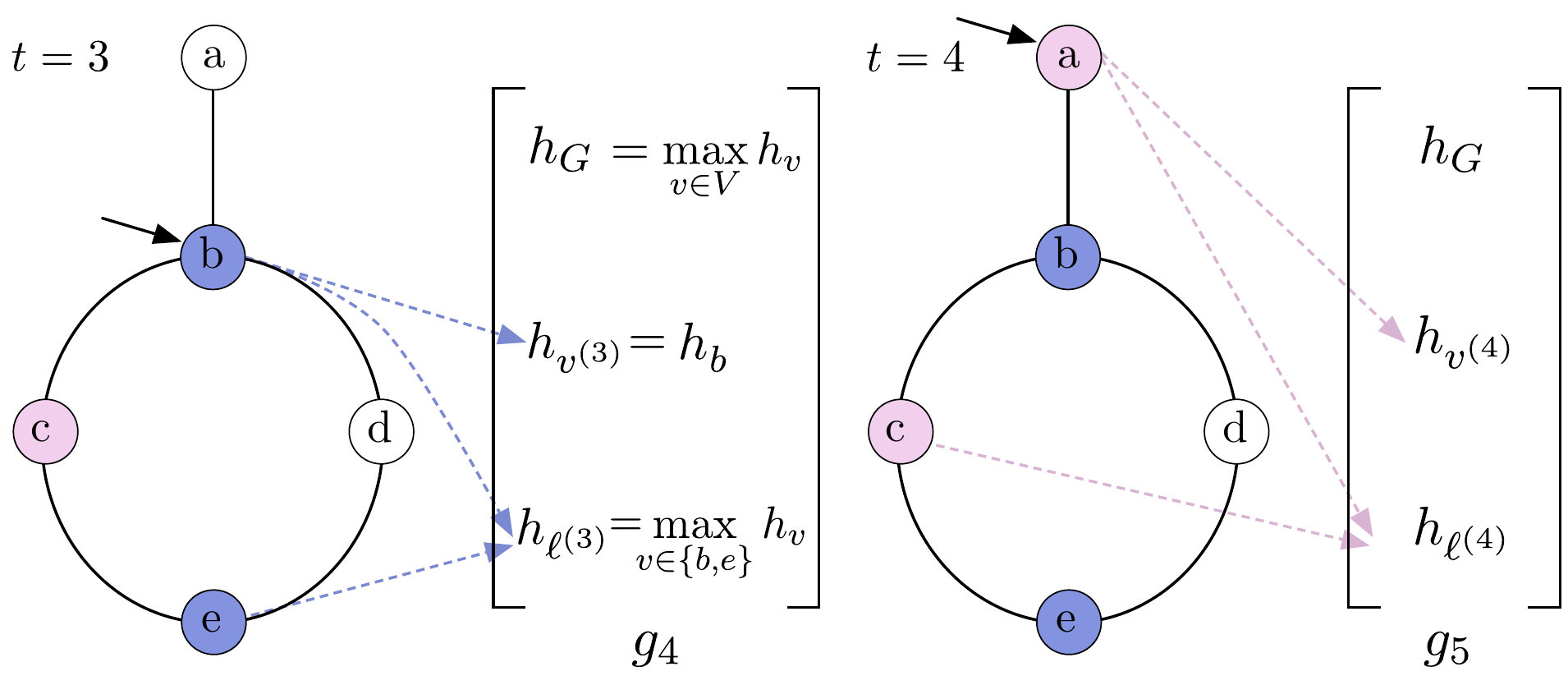}
     \caption{\emph{Temporal locality of the context embedding.} The context embedding focuses on the last labeled node. It contains the graph embedding and the embeddings of the last labeled node and its label. The example shows two states during graph coloring at time steps $t=3$ and $t=4$. At step $t=3$, the node $b$ has been colored (blue). The context embedding $g_4$ contains its embedding $h_b$ and the embedding of $b$'s label $\max (h_b, h_e)$. At step $t=4$, node $a$ is colored (pink). Now, the context embedding $g_5$ contains $h_{v^{(4)}}=h_a$ and the embedding of $a$'s label  $\max (h_a, h_c)$.} 
    \label{fig:embedding}
\end{figure}

In the first iteration we use a learned parameter $h^{(0)}$ for components (2) and (3). See \Cref{fig:embedding} for an illustration of how the context embedding changes between time steps.

\subsection{Local decoder}

The decoder takes as input the node embeddings generated by the encoder and the context embedding and outputs the next node to label.
In each time step $t$, an attention mechanism between the context embedding $g_{t}$ and each node embedding $h_v$ produces attention weights $a_v^{(t)}$. Here, we introduce a \emph{spatial locality} bias: labeling a node can only affect the attention scores of its neighbors in the next time step. %
Let $V'$ be the set of nodes already labelled. The attention weight $a^{(t)}_v$ for node $v$ in time step $t$ is given by the \emph{local decoding}. For a node $v \notin V'$:%
\begin{equation*}
    a_{v}^{(t)} = \begin{cases}
      C \cdot \tanh \left(\frac{(\Theta_1 g_t)^{T} (\Theta_2 h_{i})}{\sqrt{d}}\right)  & \text{\scriptsize{ $v \in \mathcal{N}(v^{(t-1)})$ or $t=0$}}\\ 
      a_{v}^{(t-1)} & \text{$v \notin \mathcal{N}(v^{(t-1)})$} \\
    \end{cases}
\end{equation*}
If $v \in V'$, then the attention weight is $ a_{v}^{(t)} = - \infty$. In the first iteration of the decoder, we calculate the coefficients for each node in the graph. 
As in \citet{bello2017neural}, we clip the attention coefficients within a constant range $[-C, C]$. In our experiments we set $C=10$. The learnable parameter matrices are $\Theta_1 \in \mathbb{R}^{d \times 3d}$ and $\Theta_2 \in \mathbb{R}^{d \times d}$. %
We use scaled dot-product attention~\cite{vaswani2017attention} (instead of additive attention) to speed up the decoding.
Finally, for each node $v$ we apply a softmax over all attention weights to obtain the probability $p_v$ that node $v$ is labeled next.
See Figure \ref{fig:Decoder} for a visualization of the attention weight computation during decoding. %

During inference, our \emph{greedy policy} selects the vertices with maximum probability. Our \emph{sampling policy} (for $k$ samples) runs the greedy policy once, then evaluates the learned probabilistic policy $k$ times (selecting a vertex $v$ with the learned probability $p_v$), returning the best result.

\subsection{Number of operations}

We express the number of operations (arithmetic operations and comparisons) of the model during inference parameterized by the embedding dimension $d$, the number of nodes $n$ and the number of edges $m$.
The encoder uses $\mathcal{O}(dm+d^2n)$ arithmetic operations and the decoder uses $\BigO(d^2m)$ arithmetic operations, resulting in $O(dm + d^2n + d^2 m)$ arithmetic operations, which is \emph{linear in the size of the graph}. To select the action of maximum probability (or sample a vertex), the decoder additionally needs $O(n^2)$ comparison operations (although this could be reduced). As described in \Cref{sec:experiments:scalability}, in practice the $d^2m$ term dominates the runtime for graphs up to $20,000$ vertices and the comparison operations do not dominate the computation.

\section{Experiments}\label{sec:experiments}

We evaluate our approach on established benchmarks for graph coloring and minimum vertex cover.

\textbf{Training}\; 
We use three different synthetic graph distributions to generate instances for training and validation~\cite{BarabasiAlbert_2002,Erds1984OnTE,Watts1998Collective}. 
We generate $20{,}000$ graphs for training. The graphs have between $20$ and $100$ nodes.  We use Adam with learning rate $\alpha=10^{-4}$~\cite{kingma2017adam}. The effective batch size is $B=320$, which comes from using batches of $64$ graphs for each node count $n$ and accumulating their gradients.
We clip the L2 norm of the gradient to $1$, as done in \citet{bello2017neural}. %
We selected these hyperparameters after initial experiments on the validation set.
Each model took $15-20$ CPU compute node hours to train on a cluster with Intel Xeon E5-2695 v4, 64 GB memory per node. We train each model for $200$ epochs with five random seeds and report the standard deviation $\sigma$ of cost w.r.t. the random seeds as $\pm \sigma$. %
See \Cref{apx:training} for more details.

\textbf{Test Scores}\;
In addition to mean cost, we report the ratio of the solution cost to the optimal solution cost (\emph{approximation ratio}).
For large graphs, this cannot be computed exactly in a timely manner. In this case, we use the best solution found by an ILP solver within a compute time of 1 hour. To compare with baselines which return infeasible solutions (and hence have ill-defined cost), we report the percentage of \emph{wins} (ties for first place count as wins) and the percentage of instances solved \emph{optimally}. We refer to these metric as `Wins' and `Optimal', respectively. We use the model with the lowest cost to compute these percentages.

\subsection{Results}

We compare our approach to existing greedy baselines and machine learning approaches. We focus on other heuristic approaches that return an approximation in polynomial time.

\subsubsection{Graph Coloring}

\begin{table}[t!]
\caption{Graph coloring results on the \citet{lemos} subset of the COLOR challenge graphs.}
\centering
\addtolength{\tabcolsep}{0pt}    
\begin{tabular}{@{}cllll}
\toprule & Name & \text{Cost} & \text{Wins}&  \text{Optimal} \\ \midrule
\parbox[t]{0.5mm}{\multirow{3}{*}{\rotatebox[origin=c]{90}{Classic}}} & Largest First  & $10.65$ & $50\%$    & $45\%$          \\
& DSATUR        & $9.85$ & $65\%$   & $\mathbf{50\%}$                \\
& Smallest Last & $10.8$ & $50\%$       & $45\%$        \\ \rule{0pt}{5mm}
\parbox[t]{0.5mm}{\multirow{3}{*}{\rotatebox[origin=c]{90}{ML}}} & \citet{lemos}  & N/A   & $45\%$ & $25\%$        \\ 
& Ours - Greedy         & $10.36^{\pm 0.01}$ & $55\%$ &  $\mathbf{50\%}$    \\
& Ours - Sampling       & $\mathbf{9.65}^{\pm 0.04}$ & $\mathbf{70\%}$ & $\mathbf{50\%}$     \\
\bottomrule
\end{tabular}
\addtolength{\tabcolsep}{0pt}
\label{tab:gc-results}
\end{table}

\renewcommand{\arraystretch}{1.2}

\begin{figure}[t!]
    \centering
    \vspace{1.5em}

    \subfigure[Erd\H os-R\'enyi graph.]{
		\includegraphics[width=.7\linewidth]{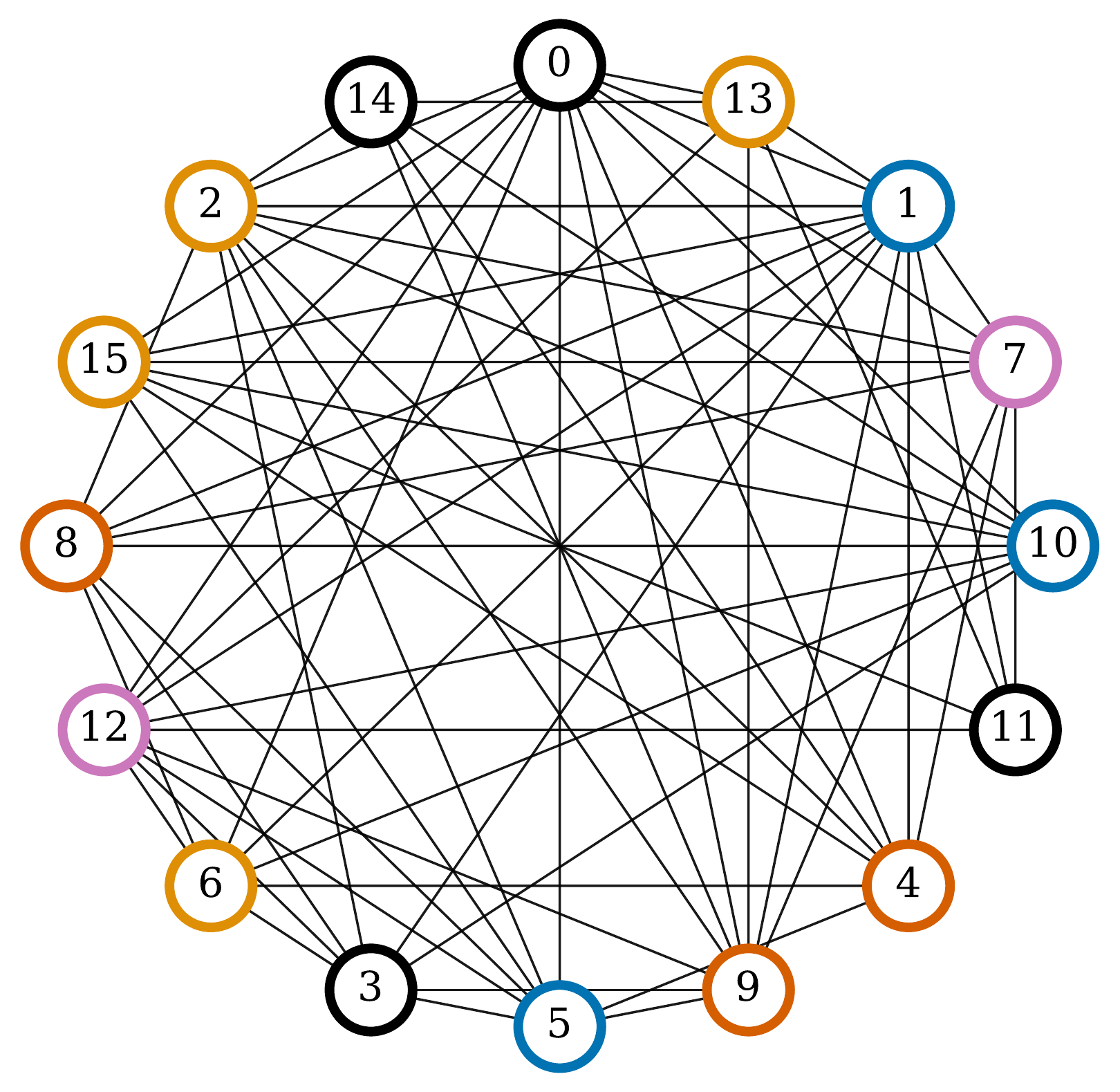} 
	}
	\subfigure[Watts-Strogatz graph.]{
		\includegraphics[width=.65\linewidth]{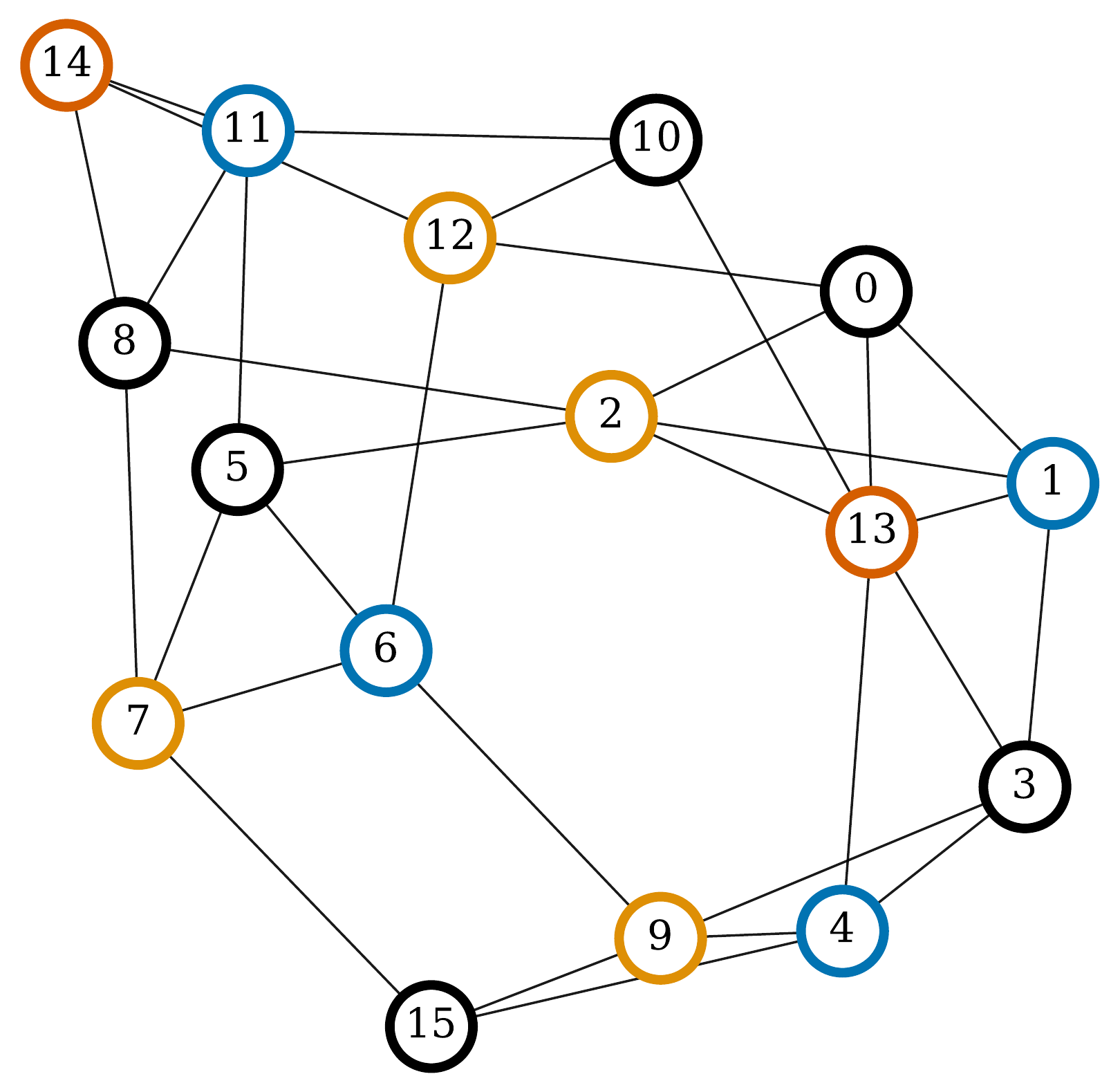} 
	}
	\subfigure[ Barab\'asi-Albert graph.]{
		\includegraphics[width=.7\linewidth]{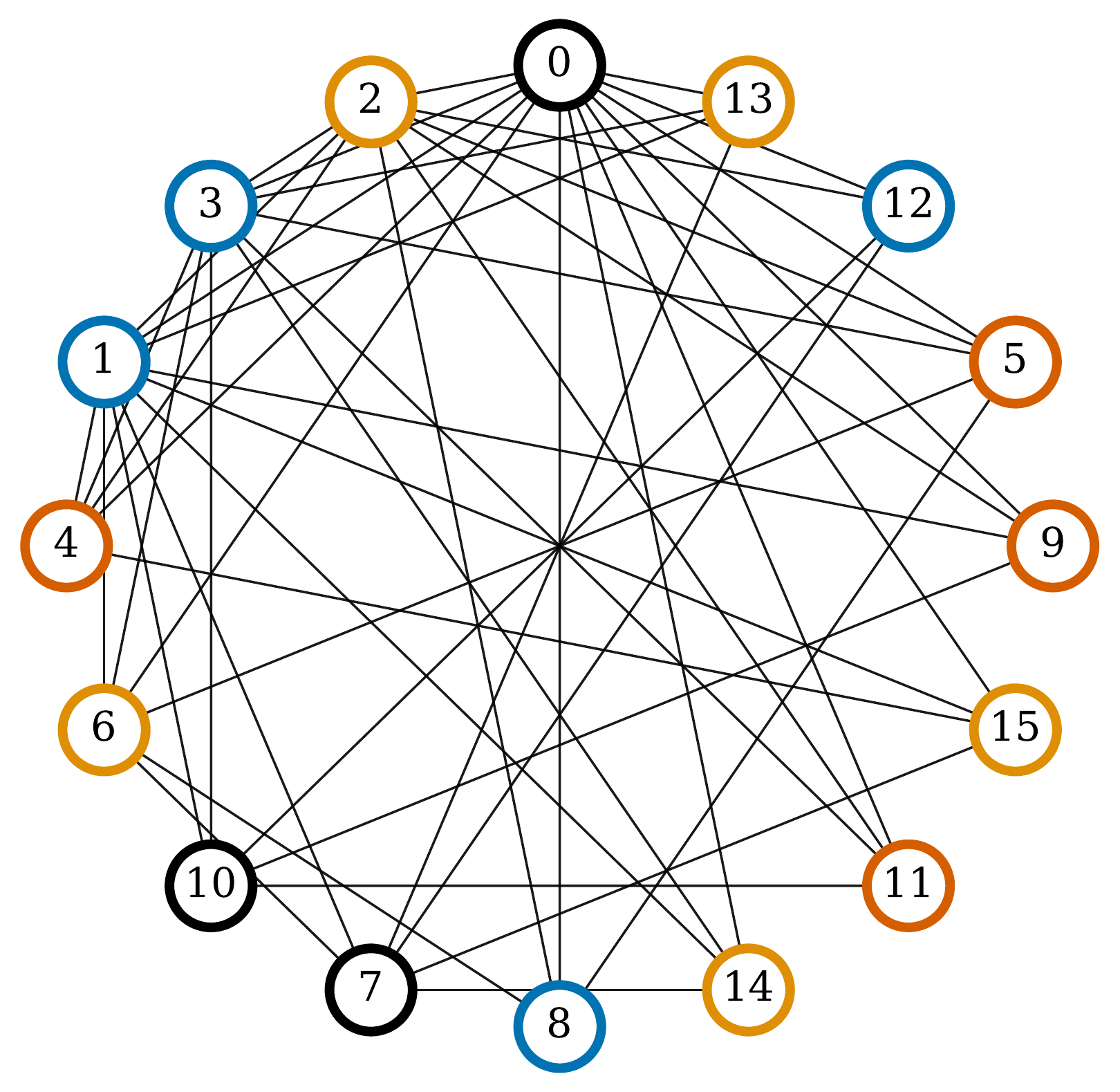} 
	}
\caption{Example colorings produced by our learned heuristic. Node borders indicate the colors. Numbers on the nodes indicate the order in which the heuristic labels them.} \label{fig:col-quali}
\end{figure}

\textbf{Greedy baselines}\; \emph{Largest-First} greedily colors nodes in decreasing order of degree. \emph{Smallest-Last}~\cite{DBLP:journals/jacm/MatulaB83} colors the nodes in reverse degeneracy order, which guarantees that when a node is colored, it will have the smallest possible number of neighbors that have been already colored. Smallest-Last guantees a constant number of colors for certain families of graphs, such as Barab\'asi-Albert graphs~\cite{BarabasiAlbert_2002} and planar graphs~\cite{DBLP:journals/jacm/MatulaB83}. \emph{DSATUR}~\cite{DSATUR} selects vertices based on the largest number of distinct colors in its neighborhood. DSATUR is exact on certain families of graphs, e.g., bipartite graphs~\cite{DSATUR}.

\textbf{Machine learning baseline}\; We compare our approach with the chromatic number estimator of \citet{lemos}. It does not guarantee that the solution is feasible, meaning that it can both under- and overestimate the chromatic number. We use the values reported by the original paper.

\textbf{Results on COLOR benchmark}\;
We evaluate our results on the same subset of the COLOR02/03 benchmark~\cite{Color02} as~\citet{lemos}, consisting of 20 instances of size between $25$ and $561$ vertices. See \Cref{tab:gc-results} for the results.

Our greedy policy outperforms both Largest-First and Smallest-Last and is tied with DSATUR for the most graphs solved optimally. When sampling ($100 $ samples) is used to evaluate the policy, our model outperforms all baselines in \emph{both mean cost and win percentage} and is also tied for the most graphs solved optimally. The approximation ratio is 1.25 and 1.13 for our greedy and sampling policies, resp.

\textbf{Qualitative Results}\;
\Cref{fig:col-quali} presents typical examples of the learned coloring heuristic on the training distribution graphs. We can observe that the heuristic generally picks higher degree, centrally located nodes first. However, if several nodes have the same degree, it favors coloring neighboring nodes subsequently. This happens in the WS graphs, see \Cref{fig:col-quali}b. 
The learned heuristic can consistently color the WS graphs with $4$ colors, which matches the Smallest-Last heuristic. We conclude that the learned heuristic captures complex aspects of the graph extending beyond simple degree-based decisions and considers the graph's local neighborhood structure.

\subsubsection{Minimum Vertex Cover}

\textbf{Greedy baselines}\; MVCApprox iteratively picks an edge that does have one of its endpoints labeled with $1$ and labels both endpoints with $1$. MVCApprox-Greedy proceeds similarly, but greedily selects the edge with maximum sum of the degrees of its endpoints. Both algorithms guarantee a $2$-approximation \cite{82PapadimitriouCombinatorial}.

\textbf{Machine learning baselines}\;
 \emph{S2V-DQN} is a $Q$-learning based approach~\cite{17Dai}. We use the values reported in the original paper. ~\citet{li18gcntreesearch} present a tree-search based approach trained in a supervised way. In contrast to S2V-DQN, it samples multiple solutions, then verifies if they are feasible. The time to construct a feasible solution varies depending on the instance. We use the publicly available code and pretrained model from the authors in our experiments. We run \citeauthor{li18gcntreesearch}'s code until it finds a feasible solution, and allow it to sample more solutions if it is below the time budget of $30$ seconds per graph.

\textbf{Results on in-distribution graphs}\;
We evaluate and compare our approach for MVC with S2V-DQN~\cite{17Dai} and~\citet{li18gcntreesearch} on the same dataset of generated graphs as.~\citet{17Dai}. It consists of 16000 graphs from two distributions, Erd\H os-R\'enyi (ER) \cite{Erds1984OnTE} and Barab\'asi-Albert (BA) \cite{BarabasiAlbert_2002}, of sizes varying from 20 to 600 nodes.
We use the results reported by \citet{17Dai} on their model trained on $40-50$ nodes, except for the graphs with less than $40$ nodes, for which no data is available for this model. Hence we use their model trained on $20-40$ nodes on these smaller graphs. %
See \Cref{tab:mvc-er-results} for the results on ER graphs and \Cref{tab:mvc-ba-results} for the results on BA graphs. %

\begin{table}[t]
\centering
\caption{Comparison of MVC approaches on dense ER graphs with edge-probability $0.15$. }

\addtolength{\tabcolsep}{2pt}    
\begin{tabular}{@{}clll@{}}
\toprule & Name & Cost & \text{Approx. Ratio}\\ \midrule
\parbox[t]{0.5mm}{\multirow{3}{*}{\rotatebox[origin=c]{90}{ML}}} & \citet{li18gcntreesearch}   & \textbf{212.296} & 1.0594 \\
& S2V-DQN    & N/A     & 1.1208  \\   
& Ours - Greedy   & 221.52$^{\pm 1.11}$ & 1.0510 \\
& Ours - 10 Samples   & 220.27$^{\pm 1.20}$& \textbf{1.0443} \\
\bottomrule
\end{tabular}
\addtolength{\tabcolsep}{2pt}
\label{tab:mvc-er-results}
\end{table}

On ER graphs, our model achieves the closest average approximation ratio, followed by \citet{li18gcntreesearch}. On the BA graphs, our model is about $2.3 \%$ away from optimal. In comparison, the two machine learning baselines are slightly less than $1\%$ away from optimal. %

\begin{table}[t]
\centering
\caption{Comparison of MVC approaches on BA graphs with average degree $4$.}

\addtolength{\tabcolsep}{+2pt}    
\begin{tabular}{@{}clll@{}}
\toprule & Name & Cost & \text{Approx. Ratio}\\ \midrule
\parbox[t]{0.5mm}{\multirow{3}{*}{\rotatebox[origin=c]{90}{ML}}} & \citet{li18gcntreesearch}   & \textbf{131.62} & \textbf{1.0084} \\
& S2V-DQN    & N/A       & 1.0099  \\   
& Ours - Greedy   & 133.78$^{\pm 0.07}$ & 1.0234\\
& Ours - 10 Samples   & 133.39${\pm 0.05}$  & 1.0202\\
\bottomrule
\end{tabular}
\addtolength{\tabcolsep}{-2pt}
\label{tab:mvc-ba-results}
\end{table}

\textbf{Results on real-world graph}\;
We evaluate our approach on the \texttt{memetracker} graph from~\citet{17Dai} which has $960$ vertices and $4{,}888$ edges. Note that the S2V-DQN model is trained on subgraphs from the same graph, giving it an advantage over the other models that have not seen the graph during training.
See \Cref{tab:mvc-memetracker-results} for the results. Our approach took less than $1$ second to find a vertex cover, whereas ~\citet{li18gcntreesearch} took $25$ minutes to find a solution on the same machine. %

\begin{table}[t]
\centering
\caption{Comparison of MVC heuristics on the \texttt{memetracker} graph. }

\addtolength{\tabcolsep}{1pt}    
\begin{tabular}{@{}clll@{}}
\toprule & Name & Cost & \text{Approx. Ratio}\\ \midrule
\parbox[t]{0.5mm}{\multirow{2}{*}{\rotatebox[origin=c]{90}{Classic}}} 
& MVCApprox        & 666   & 1.408      \\     
& MVCApprox-Greedy & 578   & 1.222               \\    \rule{0pt}{5mm}
\parbox[t]{0.5mm}{\multirow{3}{*}{\rotatebox[origin=c]{90}{ML}}} 
& \citet{li18gcntreesearch}               & 475  & 1.0042         \\
& S2V-DQN$^{\dag}$            & \textbf{474}   & \textbf{1.002}      \\
& Ours - Greedy            & \text{484$^{\pm 2.10}$}  & 1.0273   \\
\bottomrule
\end{tabular}
\addtolength{\tabcolsep}{1pt}
\\$^{\dag}$ trained and evaluated on the same graph.
\label{tab:mvc-memetracker-results}
\end{table}

\begin{figure}[t!]
    \centering

	\subfigure[Erd\H os-R\'enyi graph. ]{
		\includegraphics[width=.75\linewidth]{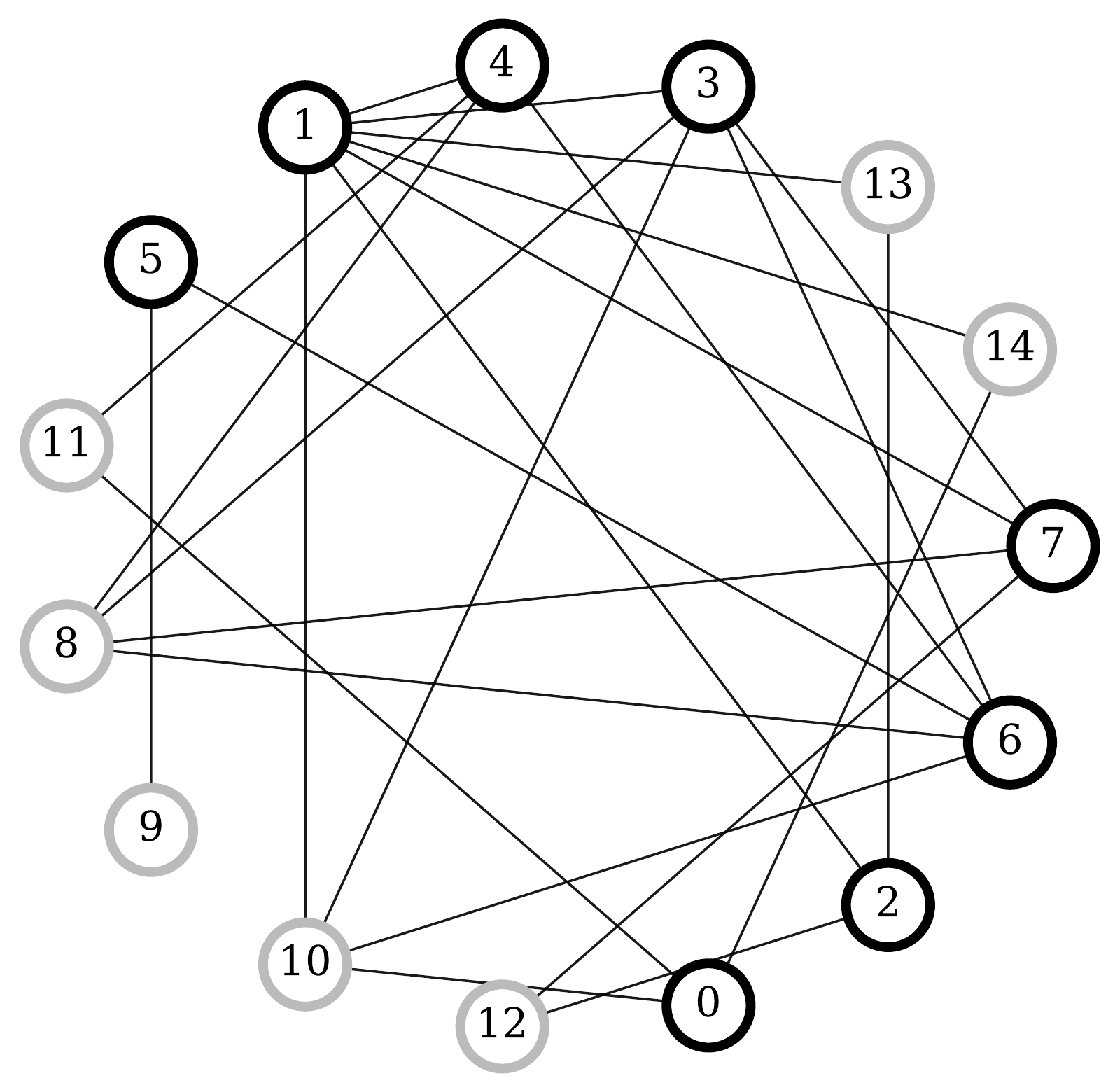}
	}
    \subfigure[Barab\'asi-Albert graph.]{
		\includegraphics[width=.75\linewidth]{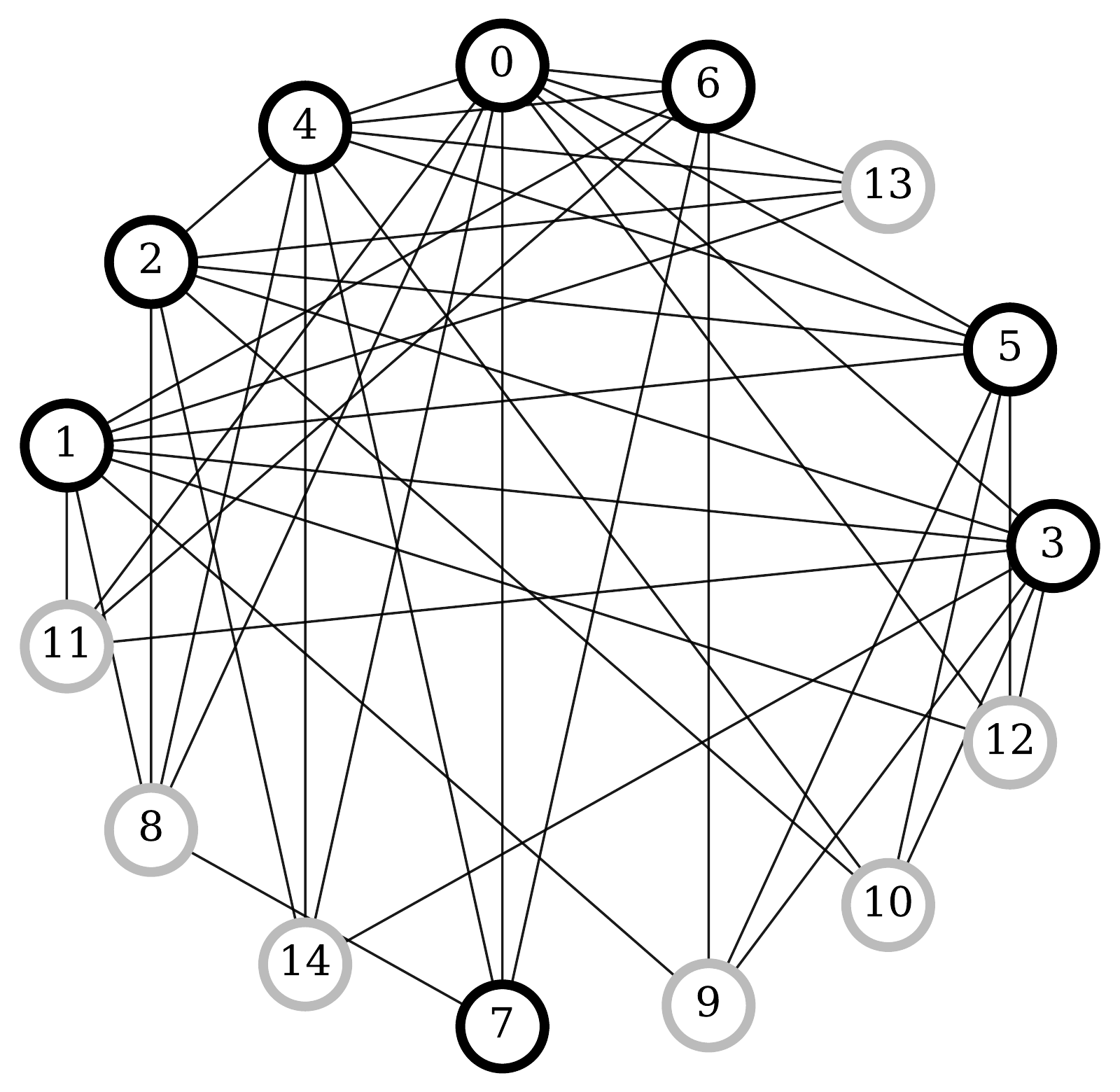} 
	}
\caption{Examples of covers produced by our learned heuristic: nodes with a bold black border are in the cover. The numbers indicate the order in which nodes are labelled. Note that as soon as a cover is found, the order of the nodes is irrelevant: they are all labeled to be outside the cover.} \label{fig:mvc-quali}
\end{figure}

\textbf{Qualitative Results};\
\Cref{fig:mvc-quali} shows typical results of our learned minimum vertex cover heuristics. On the ER graphs, we can see that the heuristic does not always start with the highest degree node. In contrast, on the BA graphs, the heuristic has a strong preference to start with the highest degree node. In contrast to the classic greedy heuristics (and our learned graph coloring heuristic), the learned MVC heuristics seldomly pick neighboring nodes subsequently. %

\subsection{Investigation of locality}\label{sec:experiments:ablation}

\subsubsection{Spatial locality}\; We test the inductive biases we made regarding locality of the decoder. First, we compare against a variant of the decoder that never updates the attention weights (called \emph{static decoding}) and a decoder that always updates all of the attention weights (called \emph{global decoding}).

\textbf{\emph{Static Decoding}} never recomputes the attention weights. For node a node $i$ that is not yet labeled, its weight is: %
\begin{equation*} \label{eq:decoding-static}
    a_{i} = %
      C \cdot \tanh \left(\frac{(\Theta_1 \mathbf{g_0} )^{T} (\Theta_2 h_{i})}{\sqrt{d}}\right)  %
\end{equation*}
Static decoding uses $O(d^2 n + m + n^2)$ operations, which are fewer than those of local update decoding when $m \gg d^2n$. With static decoding, the model is essentially a GNN with a special node-readout function. 

We train graph coloring models with static decoding. On the \citet{lemos} subset of the COLOR challenge graphs, static decoding achieves a worse mean cost of $10.74^{\pm 0.12}$ when using the greedy policy. 

\textbf{\emph{Global Decoding}}\label{sec:global-decoding} recomputes the attention weights in each time step $t$. For a node $i$ that is not yet labeled, its weight is:
\begin{equation*} \label{eq:decoding-global}
    a_{i}^{(t)} = %
      C \cdot \tanh \left(\frac{(\Theta_1 \mathbf{g_t})^{T} (\Theta_2 h_{i})}{\sqrt{d}}\right)  %
\end{equation*}
Global decoding uses $\BigO(d^2 n^2)$ operations, which is at least a $d^2$ factor more than local update decoding for not too dense graphs $(m \ll n^2 / d^2)$. When there are only two labels (as for MVC), global decoding is very similar to the \citet{kool2019attention} model. The difference to \citet{kool2019attention} is that they use additional attention layer to compute a new context embedding from $g_t$. Then, in each decoding step, they apply an attention mechanism between the context node and all the nodes that are not yet taken. %

Global decoding also achieves a worse mean cost of $10.71^{\pm 0.05}$ (greedy policy) on the \citet{lemos} graph coloring test set.

\subsubsection{Temporal locality}\; We varied the size of the context embedding (i.e., the number of nodes and their labels that contribute to it). Increasing the context size does not significantly improve the test score on graph coloring. For graph coloring, a context of size two and three results in a mean cost of $10.49^{\pm 0.12}$ and $10.42^{\pm 0.12}$, respectively, for the greedy policy.

\section{Conclusion}\label{sec:conclusion}

We presented a unifying framework for learning efficient heuristics to node labeling problems. Since such problems underly many practical applications, providing a comprehensive framework for them broadens the reach and benefits of machine learning. This work contributes to the goal of replacing hand-crafted heuristics with \emph{learned} heuristics tailored to the problems at hand. 
We demonstrated excellent results on graph coloring and minimum vertex cover as example problems.
Future work could include generalizing the architecture to handle weighted graphs and edge labeling problems.
In contrast to previous work, our work extends beyond tasks that are simpler to formulate, like minimum vertex cover, to more challenging problems like graph coloring. Formulating a new learning problem in our framework requires a cost function, a labeling rule, and an extensibility test. Our architecture benefits from two inductive biases: a spatial and a temporal bias. These biases allow our model to have nearly-linear operation number scaling and outperform several baselines.

\bibliography{refs.bib}

\begin{thebibliography}{55}
\providecommand{\natexlab}[1]{#1}
\providecommand{\url}[1]{\texttt{#1}}
\expandafter\ifx\csname urlstyle\endcsname\relax
  \providecommand{\doi}[1]{doi: #1}\else
  \providecommand{\doi}{doi: \begingroup \urlstyle{rm}\Url}\fi

\bibitem[Col(2002)]{Color02}
Computational symposium on graph coloring and generalizations ({COLOR02}),
  {I}thaca, {NY}, 7-8 {S}eptember 2002, 2002.
\newblock URL \url{https://mat.tepper.cmu.edu/COLOR02/}.

\bibitem[Abu-khzam et~al.(2004)Abu-khzam, Collins, Fellows, Langston, Suters,
  and Symons]{mvc_biology}
Abu-khzam, F., Collins, R., Fellows, M., Langston, M., Suters, W., and Symons,
  C.
\newblock Kernelization algorithms for the vertex cover problem: Theory and
  experiments.
\newblock pp.\  62--69, 01 2004.

\bibitem[Albert \& Barabási(2002)Albert and Barabási]{BarabasiAlbert_2002}
Albert, R. and Barabási, A.-L.
\newblock Statistical mechanics of complex networks.
\newblock \emph{Reviews of Modern Physics}, 74\penalty0 (1):\penalty0 47–97,
  Jan 2002.
\newblock ISSN 1539-0756.
\newblock \doi{10.1103/revmodphys.74.47}.
\newblock URL \url{http://dx.doi.org/10.1103/RevModPhys.74.47}.

\bibitem[Arora et~al.(2009)Arora, Rao, and
  Vazirani]{DBLP:journals/jacm/AroraRV09}
Arora, S., Rao, S., and Vazirani, U.~V.
\newblock Expander flows, geometric embeddings and graph partitioning.
\newblock \emph{J. {ACM}}, 56\penalty0 (2):\penalty0 5:1--5:37, 2009.
\newblock \doi{10.1145/1502793.1502794}.
\newblock URL \url{https://doi.org/10.1145/1502793.1502794}.

\bibitem[Bandh et~al.(2009)Bandh, Carle, and
  Sanneck]{DBLP:conf/iwcmc/BandhCS09}
Bandh, T., Carle, G., and Sanneck, H.
\newblock Graph coloring based physical-cell-id assignment for {LTE} networks.
\newblock In \emph{Proceedings of the International Conference on Wireless
  Communications and Mobile Computing: Connecting the World Wirelessly, {IWCMC}
  2009, Leipzig, Germany, June 21-24, 2009}, pp.\  116--120, 2009.
\newblock \doi{10.1145/1582379.1582406}.
\newblock URL \url{https://doi.org/10.1145/1582379.1582406}.

\bibitem[Barrett et~al.(2020)Barrett, Clements, Foerster, and
  Lvovsky]{barrett2019exploratory}
Barrett, T.~D., Clements, W.~R., Foerster, J.~N., and Lvovsky, A.
\newblock Exploratory combinatorial optimization with reinforcement learning.
\newblock In \emph{The Thirty-Fourth {AAAI} Conference on Artificial
  Intelligence, {AAAI} 2020, The Thirty-Second Innovative Applications of
  Artificial Intelligence Conference, {IAAI} 2020, The Tenth {AAAI} Symposium
  on Educational Advances in Artificial Intelligence, {EAAI} 2020, New York,
  NY, USA, February 7-12, 2020}, pp.\  3243--3250, 2020.
\newblock URL \url{https://aaai.org/ojs/index.php/AAAI/article/view/5723}.

\bibitem[Bello et~al.(2017)Bello, Pham, Le, Norouzi, and
  Bengio]{bello2017neural}
Bello, I., Pham, H., Le, Q.~V., Norouzi, M., and Bengio, S.
\newblock Neural combinatorial optimization with reinforcement learning.
\newblock In \emph{5th International Conference on Learning Representations,
  {ICLR} 2017, Toulon, France, April 24-26, 2017, Workshop Track Proceedings},
  2017.
\newblock URL \url{https://openreview.net/forum?id=Bk9mxlSFx}.

\bibitem[Bhattacharya et~al.(2017)Bhattacharya, Henzinger, and
  Nanongkai]{DBLP:conf/soda/BhattacharyaHN17}
Bhattacharya, S., Henzinger, M., and Nanongkai, D.
\newblock Fully dynamic approximate maximum matching and minimum vertex cover
  in \emph{O}(log\({}^{\mbox{3}}\) \emph{n}) worst case update time.
\newblock In \emph{Proceedings of the Twenty-Eighth Annual {ACM-SIAM} Symposium
  on Discrete Algorithms, {SODA} 2017, Barcelona, Spain, Hotel Porta Fira,
  January 16-19}, pp.\  470--489, 2017.
\newblock \doi{10.1137/1.9781611974782.30}.
\newblock URL \url{https://doi.org/10.1137/1.9781611974782.30}.

\bibitem[Bodlaender(2005)]{DBLP:conf/sofsem/Bodlaender05}
Bodlaender, H.~L.
\newblock Discovering treewidth.
\newblock In \emph{{SOFSEM} 2005: Theory and Practice of Computer Science, 31st
  Conference on Current Trends in Theory and Practice of Computer Science,
  Liptovsk{\'{y}} J{\'{a}}n, Slovakia, January 22-28, 2005, Proceedings}, pp.\
  1--16, 2005.
\newblock \doi{10.1007/978-3-540-30577-4\_1}.
\newblock URL \url{https://doi.org/10.1007/978-3-540-30577-4\_1}.

\bibitem[Br{\'{e}}laz(1979)]{DSATUR}
Br{\'{e}}laz, D.
\newblock New methods to color vertices of a graph.
\newblock \emph{Commun. {ACM}}, 22\penalty0 (4):\penalty0 251--256, 1979.
\newblock \doi{10.1145/359094.359101}.
\newblock URL \url{https://doi.org/10.1145/359094.359101}.

\bibitem[Cappart et~al.(2020)Cappart, Moisan, Rousseau,
  Pr{\'{e}}mont{-}Schwarz, and Cir{\'{e}}]{cappart2020combining}
Cappart, Q., Moisan, T., Rousseau, L., Pr{\'{e}}mont{-}Schwarz, I., and
  Cir{\'{e}}, A.~A.
\newblock Combining reinforcement learning and constraint programming for
  combinatorial optimization.
\newblock \emph{CoRR}, abs/2006.01610, 2020.
\newblock URL \url{https://arxiv.org/abs/2006.01610}.

\bibitem[Chaitin(1982)]{Chaitinregalloc}
Chaitin, G.~J.
\newblock Register allocation {\&} spilling via graph coloring.
\newblock In \emph{Proceedings of the {SIGPLAN} '82 Symposium on Compiler
  Construction, Boston, Massachusetts, USA, June 23-25, 1982}, pp.\  98--105,
  1982.
\newblock \doi{10.1145/800230.806984}.
\newblock URL \url{https://doi.org/10.1145/800230.806984}.

\bibitem[Cowen et~al.(1986)Cowen, Cowen, and
  Woodall]{DBLP:journals/jgt/CowenCW86}
Cowen, L.~J., Cowen, R., and Woodall, D.~R.
\newblock Defective colorings of graphs in surfaces: Partitions into subgraphs
  of bounded valency.
\newblock \emph{J. Graph Theory}, 10\penalty0 (2):\penalty0 187--195, 1986.
\newblock \doi{10.1002/jgt.3190100207}.
\newblock URL \url{https://doi.org/10.1002/jgt.3190100207}.

\bibitem[Dai et~al.(2017)Dai, Khalil, Zhang, Dilkina, and Song]{17Dai}
Dai, H., Khalil, E.~B., Zhang, Y., Dilkina, B., and Song, L.
\newblock Learning combinatorial optimization algorithms over graphs.
\newblock \emph{CoRR}, abs/1704.01665, 2017.
\newblock URL \url{http://arxiv.org/abs/1704.01665}.

\bibitem[Dantzig et~al.(1954)Dantzig, Fulkerson, and
  Johnson]{DBLP:journals/ior/DantzigFJ54}
Dantzig, G.~B., Fulkerson, D.~R., and Johnson, S.~M.
\newblock Solution of a large-scale traveling-salesman problem.
\newblock \emph{Oper. Res.}, 2\penalty0 (4):\penalty0 393--410, 1954.
\newblock \doi{10.1287/opre.2.4.393}.
\newblock URL \url{https://doi.org/10.1287/opre.2.4.393}.

\bibitem[Delbot \& Laforest(2008)Delbot and Laforest]{listright}
Delbot, F. and Laforest, C.
\newblock A better list heuristic for vertex cover.
\newblock \emph{Inf. Process. Lett.}, 107\penalty0 (3-4):\penalty0 125--127,
  2008.
\newblock \doi{10.1016/j.ipl.2008.02.004}.
\newblock URL \url{https://doi.org/10.1016/j.ipl.2008.02.004}.

\bibitem[Drori et~al.(2020)Drori, Kharkar, Sickinger, Kates, Ma, Ge, Dolev,
  Dietrich, Williamson, and Udell]{drori2020learning}
Drori, I., Kharkar, A., Sickinger, W.~R., Kates, B., Ma, Q., Ge, S., Dolev, E.,
  Dietrich, B., Williamson, D.~P., and Udell, M.
\newblock Learning to solve combinatorial optimization problems on real-world
  graphs in linear time.
\newblock In \emph{19th {IEEE} International Conference on Machine Learning and
  Applications, {ICMLA} 2020, Miami, FL, USA, December 14-17, 2020}, pp.\
  19--24, 2020.
\newblock \doi{10.1109/ICMLA51294.2020.00013}.
\newblock URL \url{https://doi.org/10.1109/ICMLA51294.2020.00013}.

\bibitem[Erd{\H{o}}s \& R{\'e}nyi(1960)Erd{\H{o}}s and
  R{\'e}nyi]{erdHos1960evolution}
Erd{\H{o}}s, P. and R{\'e}nyi, A.
\newblock On the evolution of random graphs.
\newblock \emph{Publ. Math. Inst. Hung. Acad. Sci}, 5\penalty0 (1):\penalty0
  17--60, 1960.

\bibitem[Erdős \& R{\'e}nyi(1984)Erdős and R{\'e}nyi]{Erds1984OnTE}
Erdős, P. and R{\'e}nyi, A.
\newblock On the evolution of random graphs.
\newblock \emph{Transactions of the American Mathematical Society},
  286:\penalty0 257--257, 1984.

\bibitem[Garey \& Johnson(1990)Garey and Johnson]{GareyNP-Complete}
Garey, M.~R. and Johnson, D.~S.
\newblock \emph{Computers and Intractability; A Guide to the Theory of
  NP-Completeness}.
\newblock W. H. Freeman \& Co., USA, 1990.
\newblock ISBN 0716710455.

\bibitem[Ghaffari et~al.(2020)Ghaffari, Jin, and
  Nilis]{DBLP:conf/spaa/GhaffariJN20}
Ghaffari, M., Jin, C., and Nilis, D.
\newblock A massively parallel algorithm for minimum weight vertex cover.
\newblock In \emph{{SPAA} '20: 32nd {ACM} Symposium on Parallelism in
  Algorithms and Architectures, Virtual Event, USA, July 15-17, 2020}, pp.\
  259--268, 2020.
\newblock \doi{10.1145/3350755.3400260}.
\newblock URL \url{https://doi.org/10.1145/3350755.3400260}.

\bibitem[Hagberg et~al.(2008)Hagberg, Schult, and Swart]{networkx}
Hagberg, A.~A., Schult, D.~A., and Swart, P.~J.
\newblock Exploring network structure, dynamics, and function using networkx.
\newblock In Varoquaux, G., Vaught, T., and Millman, J. (eds.),
  \emph{Proceedings of the 7th Python in Science Conference}, pp.\  11 -- 15,
  Pasadena, CA USA, 2008.

\bibitem[Harary \& Melter(1976)Harary and Melter]{harary1976metric}
Harary, F. and Melter, R.~A.
\newblock On the metric dimension of a graph.
\newblock \emph{Ars Combinatoria}, 2\penalty0 (191-195):\penalty0 1, 1976.

\bibitem[He et~al.(2016)He, Zhang, Ren, and Sun]{skip_connections}
He, K., Zhang, X., Ren, S., and Sun, J.
\newblock Deep residual learning for image recognition.
\newblock In \emph{2016 {IEEE} Conference on Computer Vision and Pattern
  Recognition, {CVPR} 2016, Las Vegas, NV, USA, June 27-30, 2016}, pp.\
  770--778, 2016.
\newblock \doi{10.1109/CVPR.2016.90}.
\newblock URL \url{https://doi.org/10.1109/CVPR.2016.90}.

\bibitem[Hedetniemi \& Laskar(1990)Hedetniemi and
  Laskar]{DBLP:journals/dm/HedetniemiL90a}
Hedetniemi, S.~T. and Laskar, R.~C.
\newblock Bibliography on domination in graphs and some basic definitions of
  domination parameters.
\newblock \emph{Discret. Math.}, 86\penalty0 (1-3):\penalty0 257--277, 1990.
\newblock \doi{10.1016/0012-365X(90)90365-O}.
\newblock URL \url{https://doi.org/10.1016/0012-365X(90)90365-O}.

\bibitem[Huang et~al.(2019)Huang, Patwary, and
  Diamos]{DBLP:journals/corr/abs-1902-10162}
Huang, J., Patwary, M. M.~A., and Diamos, G.~F.
\newblock Coloring big graphs with alphagozero.
\newblock \emph{CoRR}, abs/1902.10162, 2019.
\newblock URL \url{http://arxiv.org/abs/1902.10162}.

\bibitem[Ioffe \& Szegedy(2015)Ioffe and Szegedy]{batch_normalization}
Ioffe, S. and Szegedy, C.
\newblock Batch normalization: Accelerating deep network training by reducing
  internal covariate shift.
\newblock In \emph{Proceedings of the 32nd International Conference on Machine
  Learning, {ICML} 2015, Lille, France, 6-11 July 2015}, pp.\  448--456, 2015.
\newblock URL \url{http://proceedings.mlr.press/v37/ioffe15.html}.

\bibitem[Jensen et~al.(1995)Jensen, Jensen, and Toft]{jensen1995graph}
Jensen, T., Jensen, T., and Toft, B.
\newblock \emph{Graph Coloring Problems}.
\newblock A Wiley interscience publication. Wiley, 1995.
\newblock ISBN 9780471028659.
\newblock URL \url{https://books.google.ch/books?id=YfZQAAAAMAAJ}.

\bibitem[Joshi et~al.(2019)Joshi, Laurent, and Bresson]{joshi}
Joshi, C.~K., Laurent, T., and Bresson, X.
\newblock An efficient graph convolutional network technique for the travelling
  salesman problem.
\newblock \emph{CoRR}, abs/1906.01227, 2019.
\newblock URL \url{http://arxiv.org/abs/1906.01227}.

\bibitem[Karger \& Stein(1996)Karger and Stein]{DBLP:journals/jacm/KargerS96}
Karger, D.~R. and Stein, C.
\newblock A new approach to the minimum cut problem.
\newblock \emph{J. {ACM}}, 43\penalty0 (4):\penalty0 601--640, 1996.
\newblock \doi{10.1145/234533.234534}.
\newblock URL \url{https://doi.org/10.1145/234533.234534}.

\bibitem[Karger et~al.(1997)Karger, Motwani, and
  Ramkumar]{DBLP:journals/algorithmica/KargerMR97}
Karger, D.~R., Motwani, R., and Ramkumar, G. D.~S.
\newblock On approximating the longest path in a graph.
\newblock \emph{Algorithmica}, 18\penalty0 (1):\penalty0 82--98, 1997.
\newblock \doi{10.1007/BF02523689}.
\newblock URL \url{https://doi.org/10.1007/BF02523689}.

\bibitem[Karp(1972)]{Kar72}
Karp, R.
\newblock Reducibility among combinatorial problems.
\newblock In Miller, R. and Thatcher, J. (eds.), \emph{Complexity of Computer
  Computations}, pp.\  85--103. Plenum Press, 1972.

\bibitem[Kernighan \& Lin(1970)Kernighan and
  Lin]{DBLP:journals/bstj/KernighanL70}
Kernighan, B.~W. and Lin, S.
\newblock An efficient heuristic procedure for partitioning graphs.
\newblock \emph{Bell Syst. Tech. J.}, 49\penalty0 (2):\penalty0 291--307, 1970.
\newblock \doi{10.1002/j.1538-7305.1970.tb01770.x}.
\newblock URL \url{https://doi.org/10.1002/j.1538-7305.1970.tb01770.x}.

\bibitem[Kingma \& Ba(2015)Kingma and Ba]{kingma2017adam}
Kingma, D.~P. and Ba, J.
\newblock Adam: {A} method for stochastic optimization.
\newblock In \emph{3rd International Conference on Learning Representations,
  {ICLR} 2015, San Diego, CA, USA, May 7-9, 2015, Conference Track
  Proceedings}, 2015.
\newblock URL \url{http://arxiv.org/abs/1412.6980}.

\bibitem[Kool et~al.(2019)Kool, van Hoof, and Welling]{kool2019attention}
Kool, W., van Hoof, H., and Welling, M.
\newblock Attention, learn to solve routing problems!
\newblock In \emph{7th International Conference on Learning Representations,
  {ICLR} 2019, New Orleans, LA, USA, May 6-9, 2019}, 2019.

\bibitem[Lee et~al.(2019)Lee, Rossi, Kim, Ahmed, and Koh]{lee2018attention}
Lee, J.~B., Rossi, R.~A., Kim, S., Ahmed, N.~K., and Koh, E.
\newblock Attention models in graphs: {A} survey.
\newblock \emph{{ACM} Trans. Knowl. Discov. Data}, 13\penalty0 (6):\penalty0
  62:1--62:25, 2019.
\newblock \doi{10.1145/3363574}.
\newblock URL \url{https://doi.org/10.1145/3363574}.

\bibitem[Lemos et~al.(2019)Lemos, Prates, Avelar, and Lamb]{lemos}
Lemos, H., Prates, M. O.~R., Avelar, P. H.~C., and Lamb, L.~C.
\newblock Graph colouring meets deep learning: Effective graph neural network
  models for combinatorial problems.
\newblock \emph{CoRR}, abs/1903.04598, 2019.
\newblock URL \url{http://arxiv.org/abs/1903.04598}.

\bibitem[Li et~al.(2018)Li, Chen, and Koltun]{li18gcntreesearch}
Li, Z., Chen, Q., and Koltun, V.
\newblock Combinatorial optimization with graph convolutional networks and
  guided tree search.
\newblock In \emph{Advances in Neural Information Processing Systems 31: Annual
  Conference on Neural Information Processing Systems 2018, NeurIPS 2018,
  December 3-8, 2018, Montr{\'{e}}al, Canada}, pp.\  537--546, 2018.
\newblock URL
  \url{https://proceedings.neurips.cc/paper/2018/hash/8d3bba7425e7c98c50f52ca1b52d3735-Abstract.html}.

\bibitem[Ma et~al.(2020)Ma, Ge, He, Thaker, and Drori]{ma2019combinatorial}
Ma, Q., Ge, S., He, D., Thaker, D., and Drori, I.
\newblock Combinatorial optimization by graph pointer networks and hierarchical
  reinforcement learning.
\newblock In \emph{AAAI Workshop on Deep Learning on Graphs: Methodologies and
  Applications}, 2020.

\bibitem[Maas et~al.(2013)Maas, Hannun, and Ng]{maas2013rectifier}
Maas, A.~L., Hannun, A.~Y., and Ng, A.~Y.
\newblock Rectifier nonlinearities improve neural network acoustic models.
\newblock In \emph{Proc. icml}, volume~30, pp.\ ~3. Citeseer, 2013.

\bibitem[Manchanda et~al.(2020)Manchanda, Mittal, Dhawan, Medya, Ranu, and
  Singh]{DBLP:conf/nips/ManchandaMDMRS20}
Manchanda, S., Mittal, A., Dhawan, A., Medya, S., Ranu, S., and Singh, A.
\newblock {GCOMB:} learning budget-constrained combinatorial algorithms over
  billion-sized graphs.
\newblock In \emph{Advances in Neural Information Processing Systems 33: Annual
  Conference on Neural Information Processing Systems 2020, NeurIPS 2020,
  December 6-12, 2020, virtual}, 2020.
\newblock URL
  \url{https://proceedings.neurips.cc/paper/2020/hash/e7532dbeff7ef901f2e70daacb3f452d-Abstract.html}.

\bibitem[Marx(2004)]{Marx03graphcolouring}
Marx, D.
\newblock Graph colouring problems and their applications in scheduling.
\newblock \emph{Periodica Polytechnica Electrical Engineering}, 48:\penalty0
  11--16, 2004.

\bibitem[Matula \& Beck(1983)Matula and Beck]{DBLP:journals/jacm/MatulaB83}
Matula, D.~W. and Beck, L.~L.
\newblock Smallest-last ordering and clustering and graph coloring algorithms.
\newblock \emph{J. {ACM}}, 30\penalty0 (3):\penalty0 417--427, 1983.
\newblock \doi{10.1145/2402.322385}.
\newblock URL \url{https://doi.org/10.1145/2402.322385}.

\bibitem[Myszkowski(2008)]{DBLP:series/sci/Myszkowski08}
Myszkowski, P.~B.
\newblock Solving scheduling problems by evolutionary algorithms for graph
  coloring problem.
\newblock In \emph{Metaheuristics for Scheduling in Industrial and
  Manufacturing Applications}, pp.\  145--167. 2008.
\newblock \doi{10.1007/978-3-540-78985-7\_7}.
\newblock URL \url{https://doi.org/10.1007/978-3-540-78985-7\_7}.

\bibitem[Onak et~al.(2012)Onak, Ron, Rosen, and
  Rubinfeld]{DBLP:conf/soda/OnakRRR12}
Onak, K., Ron, D., Rosen, M., and Rubinfeld, R.
\newblock A near-optimal sublinear-time algorithm for approximating the minimum
  vertex cover size.
\newblock In \emph{Proceedings of the Twenty-Third Annual {ACM-SIAM} Symposium
  on Discrete Algorithms, {SODA} 2012, Kyoto, Japan, January 17-19, 2012}, pp.\
   1123--1131, 2012.
\newblock \doi{10.1137/1.9781611973099.88}.
\newblock URL \url{https://doi.org/10.1137/1.9781611973099.88}.

\bibitem[Papadimitriou \& Steiglitz(1982)Papadimitriou and
  Steiglitz]{82PapadimitriouCombinatorial}
Papadimitriou, C. and Steiglitz, K.
\newblock \emph{Combinatorial Optimization: Algorithms and Complexity},
  volume~32.
\newblock 01 1982.
\newblock ISBN 0-13-152462-3.
\newblock \doi{10.1109/TASSP.1984.1164450}.

\bibitem[Robson(1986)]{DBLP:journals/jal/Robson86}
Robson, J.~M.
\newblock Algorithms for maximum independent sets.
\newblock \emph{J. Algorithms}, 7\penalty0 (3):\penalty0 425--440, 1986.
\newblock \doi{10.1016/0196-6774(86)90032-5}.
\newblock URL \url{https://doi.org/10.1016/0196-6774(86)90032-5}.

\bibitem[Smith et~al.(2004)Smith, Ramsey, and
  Holloway]{DBLP:conf/pldi/SmithRH04}
Smith, M.~D., Ramsey, N., and Holloway, G.~H.
\newblock A generalized algorithm for graph-coloring register allocation.
\newblock In \emph{Proceedings of the {ACM} {SIGPLAN} 2004 Conference on
  Programming Language Design and Implementation 2004, Washington, DC, USA,
  June 9-11, 2004}, pp.\  277--288, 2004.
\newblock \doi{10.1145/996841.996875}.
\newblock URL \url{https://doi.org/10.1145/996841.996875}.

\bibitem[Sutton \& Barto(2018)Sutton and Barto]{sutton}
Sutton, R.~S. and Barto, A.~G.
\newblock \emph{Reinforcement Learning: An Introduction}.
\newblock A Bradford Book, Cambridge, MA, USA, 2018.
\newblock ISBN 0262039249.

\bibitem[Tarjan \& Trojanowski(1977)Tarjan and
  Trojanowski]{DBLP:journals/siamcomp/TarjanT77}
Tarjan, R.~E. and Trojanowski, A.~E.
\newblock Finding a maximum independent set.
\newblock \emph{{SIAM} J. Comput.}, 6\penalty0 (3):\penalty0 537--546, 1977.
\newblock \doi{10.1137/0206038}.
\newblock URL \url{https://doi.org/10.1137/0206038}.

\bibitem[Tomita \& Seki(2003)Tomita and Seki]{DBLP:conf/dmtcs/TomitaS03}
Tomita, E. and Seki, T.
\newblock An efficient branch-and-bound algorithm for finding a maximum clique.
\newblock In \emph{Discrete Mathematics and Theoretical Computer Science, 4th
  International Conference, {DMTCS} 2003, Dijon, France, July 7-12, 2003.
  Proceedings}, pp.\  278--289, 2003.
\newblock \doi{10.1007/3-540-45066-1\_22}.
\newblock URL \url{https://doi.org/10.1007/3-540-45066-1\_22}.

\bibitem[ul~Islam \& Kalita(2017)ul~Islam and Kalita]{Islam2017ApplicationOM}
ul~Islam, A. and Kalita, B.
\newblock Application of minimum vertex cover for keyword – based text
  summarization process.
\newblock 2017.

\bibitem[Vaswani et~al.(2017)Vaswani, Shazeer, Parmar, Uszkoreit, Jones, Gomez,
  Kaiser, and Polosukhin]{vaswani2017attention}
Vaswani, A., Shazeer, N., Parmar, N., Uszkoreit, J., Jones, L., Gomez, A.~N.,
  Kaiser, L., and Polosukhin, I.
\newblock Attention is all you need.
\newblock In \emph{Advances in Neural Information Processing Systems 30: Annual
  Conference on Neural Information Processing Systems 2017, December 4-9, 2017,
  Long Beach, CA, {USA}}, pp.\  5998--6008, 2017.
\newblock URL
  \url{https://proceedings.neurips.cc/paper/2017/hash/3f5ee243547dee91fbd053c1c4a845aa-Abstract.html}.

\bibitem[Velickovic et~al.(2018)Velickovic, Cucurull, Casanova, Romero,
  Li{\`{o}}, and Bengio]{DBLP:conf/iclr/VelickovicCCRLB18}
Velickovic, P., Cucurull, G., Casanova, A., Romero, A., Li{\`{o}}, P., and
  Bengio, Y.
\newblock Graph attention networks.
\newblock In \emph{6th International Conference on Learning Representations,
  {ICLR} 2018, Vancouver, BC, Canada, April 30 - May 3, 2018, Conference Track
  Proceedings}, 2018.
\newblock URL \url{https://openreview.net/forum?id=rJXMpikCZ}.

\bibitem[Watts \& Strogatz(1998)Watts and Strogatz]{Watts1998Collective}
Watts, D.~J. and Strogatz, S.~H.
\newblock {Collective dynamics of 'small-world' networks}.
\newblock \emph{Nature}, 393\penalty0 (6684):\penalty0 440--442, June 1998.
\newblock ISSN 0028-0836.
\newblock \doi{10.1038/30918}.
\newblock URL \url{http://dx.doi.org/10.1038/30918}.

\end{thebibliography}
\bibliographystyle{icml2022}

\newpage
\appendix
\onecolumn

\section{Training}\label{apx:training}

\subsection{Data Generation} \label{sec:Data_Generation}
We use four different synthetic graph distributions to generate instances for training and validation. All graphs are generated via the Python \textsc{Networkx} library \cite{networkx}. %

\begin{description}[style=unboxed,leftmargin=0cm]
\item[Barab\'asi-Albert Model \cite{BarabasiAlbert_2002}]
The Barab\'asi-Albert (BA) Model generates random scale-free networks. Similar to real-world networks BA graphs grow by preferential attachment, i.e., a new node is more likely to link to more connected nodes. The model is parameterized by one parameter $\delta$, which dictates the average degree.

\item[Erd\H os-R\'enyi Model \cite{Erds1984OnTE}]
An Erd\H os-R\'enyi (ER) graph $G(n,p)$ has $n$ nodes and each edge exists independently with probability $p$. The expected number of edges is ${n\choose2} p$.

\item[Watts-Strogatz Model \cite{Watts1998Collective}]
Watts-Strogatz (WS) graphs were developed to overcome the shortcomings of ER graphs when modeling real world graphs. In real networks we see the formation of local clusters, i.e., the neighbors of a node are more likely to be neighbors. For parameters $k$ and $q$, a WS graph is built as follows: build a ring of $n$ nodes. Next, connect each node to its $k$ nearest neighbors. Finally, replace each edge $\{u,v\}$ by a new edge $\{u,w\}$ (chosen uniformly at random) with probability $q$.

\end{description}

\subsubsection{Training set parameters}

See \Cref{tab:model_parameters} for the parameters of the graph distributions used during training. Note that for BA and ER graphs, the parameters match those used in the \citet{17Dai} test set (see \Cref{tab:mvc-er-results} and \Cref{tab:mvc-ba-results}).  We also consider sparse ER graphs (S-ER), for we set the edge probability such that graphs have expected average degree close to $\Delta=7.5$ when $n$ is small but remain connected with high probability when $n$ is large. This means that 
\begin{align}
	p_{s-er}= \min\left(1, \max\left( \frac{\Delta}{n}, (1+\epsilon)  \frac{\ln n}{n}\right)\right) \enspace , \label{eqn:er-p}
\end{align}
for a small $\epsilon$, which we set to $0.2$ in our experiments. The formula is derived from the connectivity threshold of ER graphs~\cite{erdHos1960evolution}.

For graph coloring, we train on a hybrid dataset consisting of an equal proportion of BA, S-ER, and WS graphs. For minimum vertex cut, we train on a dataset consisting of BA graphs, a dataset consisting of ER graphs, and a hybrid dataset consisting on a combination of the two (in equal proportion). We use the in-distribution models for the evaluation on the synthetic test instances and the hybrid model for the \texttt{memetracker} graph.
During training, we use an equal proportion of graphs with $n\in \{20, 40, 50, 70, 100\}$ nodes.

\newcommand{\STAB}[1]{\begin{tabular}{@{}c@{}}#1\end{tabular}}

\begin{table}[t]
\caption{The graph parameters for training and validation.}

\centering
\addtolength{\tabcolsep}{+6pt}
\begin{tabular}{@{}lllll@{}}
\toprule
BA     & ER   & S-ER   & WS     \\ \midrule
$\delta=4$ & $p=0.15$ & $p=p_{s-er}$ & $k=5$,  \\
 & 				 &  &  $q=0.1$ \\ \bottomrule
\end{tabular}
\addtolength{\tabcolsep}{-6pt}    
\label{tab:model_parameters}
\end{table}

\subsection{Policy Optimization}\label{apx:training-policy}

We train our model with \textsc{Reinforce} with a greedy rollout baseline \citet{kool2019attention}. The details follow.
We denote the cost of labeling the graph $G_i$ in the order given by the sequence of nodes $\pi$ by $\mathcal{L}(\pi, G_i)$. A model $M$ is parameterized by parameters $\theta$. On a graph $G_i$, the model returns a sequence of nodes $\pi$ and an associated probability $p_\theta$. The probability $p_\theta$ is the product of all action probabilities that led to the sequence of nodes $\pi$. We write $p_{\theta}, \pi \gets M_{\theta}(G_i)$ when the policy is evaluated deterministically and $p_{\theta}, \pi \sim M_{\theta}(G_i)$ when the policy is evaluated probabilistically.

The complete training procedure is given in Algorithm \ref{Algorithm:policy-training}. 
Note that Algorithm \ref{Algorithm:policy-training} follows from the textbook \textsc{Reinforce} with a baseline~\cite{sutton} by factoring the probability of reaching a terminal state and using that the rewards are $0$ in our MDP except when reaching a terminal state. Unlike \citet{kool2019attention}, we do not use warmup epochs where an exponential moving average baseline is used in the first epochs of training.

\newcommand{\ALGCOMMENT}[1]{\hfill \text{#1}}

\begin{algorithm*}[t]
\caption{Policy Training with \textsc{Reinforce} + Baseline.}\label{Algorithm:policy-training}
\begin{algorithmic} 
\setstretch{1.1}
\STATE \textbf{Input:} \text{number of epochs $E$, batch size $B$, dataset $D_{\text{train}}$, earing rate $\alpha$}
\STATE \emph{Initialize model $M_\theta$ and baseline model $M_\theta^{BL}$ }
\STATE $D_{\text{challenge}} \gets$ \emph{Sample new challenge dataset}
\FOR {$\text{\emph{epoch}} = 1, \dots, E$ }
  \FOR {$\text{\emph{batch}}$ in $D_{train}$}
    \STATE $[\enspace p_{\theta, i}, \pi_i \sim  M_{\theta}(G_i) \enspace\text{for } G_i \text{ in } \text{\emph{batch}} \enspace]$ \ALGCOMMENT{// Sample from policy}

    \STATE  $[\enspace p_{\theta, i}^{BL}, \pi_i^{BL} \gets M_{\theta^{BL}}(G_i) \enspace \text{for } G_i \text{ in } \text{\emph{batch}} \enspace]$  \ALGCOMMENT{// Greedy baseline}

    \STATE { $\nabla_{\theta}J(\theta) = \frac{1}{B}  
    \sum_{i=1}^{B} (\mathcal{L}(\pi_i \mid G_i) - \mathcal{L}(\pi_i^{BL} \mid G_i)) \enspace \nabla_{\theta} \log (p_{\theta,i} )$ } \ALGCOMMENT{// Policy gradient}
    \STATE  {\normalsize $\theta \gets \emph{Gradient Descent}(\theta, \nabla_{\theta}J(\theta), \alpha)$}
  \ENDFOR
 
  \ALGCOMMENT{//Challenge the baseline} 
  \IF  {\emph{OneSidedPairedTTest}($M_\theta$, $M_\theta^{BL}, D_{\text{challenge}} )$ $< 0.05$} 
    \STATE $\theta^{BL} \gets \theta$
    \STATE $D_{\text{challenge}} \gets$ \emph{Sample new challenge dataset}
    
  \ENDIF
  
\ENDFOR

\end{algorithmic}

\end{algorithm*}

\section{Additional Results}\label{apx:experiments}

\subsection{Results by graph size}

Our test and validation data already include graphs larger than those seen in training. In the main paper (Tables 1-3), we report the average across the graph sizes. See \Cref{tab:mvc} for a breakdown of Table 2, where we show how the approximation ratio varies with the test instance size on minimum vertex cover (MVC). For graph coloring (GC), about a third of the test instances have more vertices than seen in training. The average approximation ratio on those instances is 1.18 and 1.14 for the Greedy and the Sampling policies, respectively. This is comparable to the results on the overall test set (1.25 and 1.13 for the Greedy and the Sample policy, respectively).

\setlength\tabcolsep{3pt}

\begin{table}
\caption{Cost and approximation ratio of our approach for MVC on ER graphs (10 samples) from Table 2 by instance size.}\label{tab:mvc}
\begin{tabular}{llllll}
\toprule
Vertices          & 15-20 & 40-100 & 100-300 & 300-500 & 500-600 \\
Cost       & 8     & 43.8  & 178.4   & 384.9  & 540     \\
\small{Appr.} Ratio & 1.012 & 1.042 & 1.048  & 1.055 & 1.054   \\
\bottomrule
\end{tabular}
\end{table}

\subsection{Validation Results}

We compare the cost of the learned heuristic for different parameters of the training. The validation set consists of $600$ graphs with $n$ nodes for $n\in \{20, 50, 100, 200, 400, 600\}$.%

\subsubsection{Graph coloring}

\Cref{tab:gc-validation} shows the validation cost on the three training distribution for the configuration used in the experiments. Over all three distributions, the mean validation cost is $4.95^{\pm 0.02}$.

\begin{table}
\caption{Validation Cost for GC.}
\addtolength{\tabcolsep}{0pt}    
\begin{tabular}{llll}
\toprule Training Distr. & Cost S-ER & Cost WS & Cost BA \\ \midrule
  S-ER+WS+BA & 5.32$^{\pm 0.05}$ & 4.01$^{\pm 0.00}$ & 5.50$^{\pm 0.04}$ \\
\bottomrule
\end{tabular}
\addtolength{\tabcolsep}{-0pt}
\label{tab:gc-validation}
\end{table}

\textbf{Number of attention heads}
We varied the number of attention heads (among $1,2,4$) while keeping the dimension per head to $16$. For graph coloring, this results in a mean validation cost of $5.29^{\pm 0.02}$, $4.98^{\pm 0.01}$, and $4.95^{\pm 0.02}$, respectively. Hence, $4$ attention heads (overall hidden dimension $64$) yields the lowest mean validation cost for graph coloring.

\textbf{Learning rate}
With a larger learning rate of $\alpha=10^{-3}$, the mean validation cost for graph coloring is significantly worse, namely $5.22^{\pm 0.002}$. A smaller learning rate of $\alpha=10^{-5}$ leads to a mean validation cost of $5.02^{\pm 0.001}$, which is slightly worse than the cost for $\alpha=10^{-4}$.

\subsubsection{Minimum vertex cover}

\Cref{tab:mvc-validation} shows the validation results for training on either only one distribution and evaluating on ER and BA graphs. Training on a mixture ER and BA graphs leads to worse validation cost on BA graphs compared to training only on BA graphs. Training on ER graphs exclusively without BA graphs leads to a slight cost improvement on ER graphs.

\begin{table}
\caption{Validation Cost for MVC.}
\addtolength{\tabcolsep}{1pt}    
\begin{tabular}{lll}
\toprule Training Distribution & Cost ER & Cost BA \\ \midrule
  ER & \textbf{223.56}$^{\pm 0.11}$ & 218.07$^{\pm 4.36}$ \\
 BA   &  223.76$^{\pm 0.43}$ & \textbf{199.44}$^{\pm 12.02}$  \\   
 ER+BA  & 223.98$^{\pm 0.26}$ &  208.18$^{\pm 6.24}$ \\
\bottomrule
\end{tabular}
\addtolength{\tabcolsep}{-1pt}
\label{tab:mvc-validation}
\end{table}

\subsection{Runtime Scalability}\label{sec:experiments:scalability}

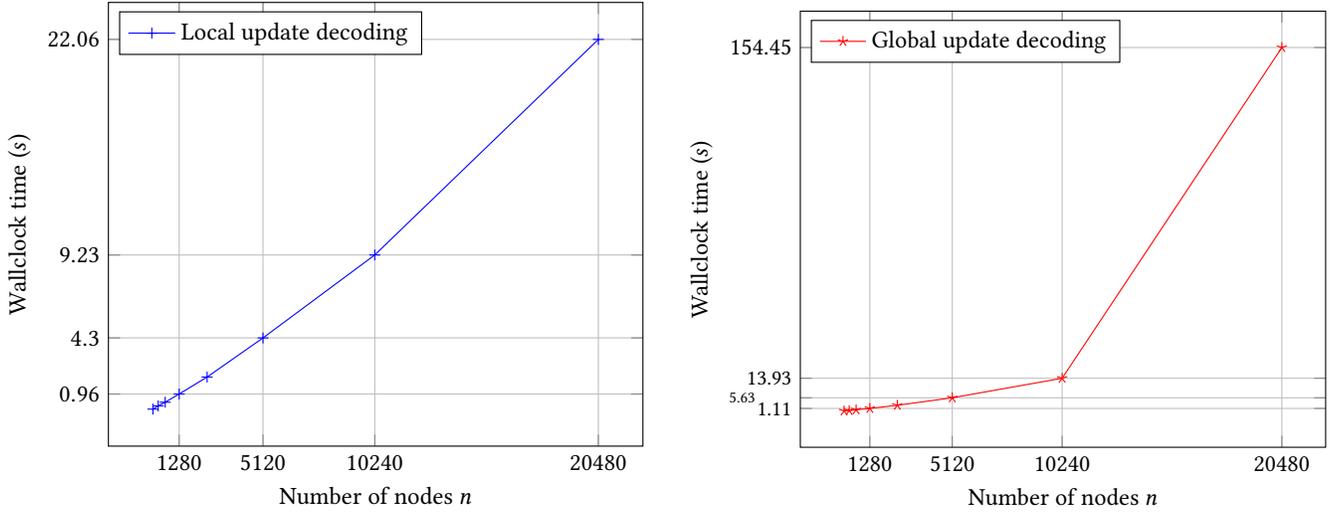
\begin{figure*}[t!]

\resizebox{0.49\linewidth}{!} {
\begin{tikzpicture}\begin{axis}[
title={}, xlabel={Number of nodes $n$}, ylabel={Wallclock time ($s$)}, grid=major,
legend entries={Local update decoding},
xtick={1280, 5120, 10240, 20480},
xticklabels={$1280$, $5120$, $10240$, $20480$},
xtick scale label code/.code={},
ytick={0.96,4.3,9.23,22.06},
legend style={at={(0.02,0.98)},anchor=north west}
]
\addplot[blue,mark=+] table [x=n, y=t-gc-local, col sep=comma] {./Data/Runtime/Runtime-gc-local.csv};
\end{axis}
\end{tikzpicture}
}
\hfill
\resizebox{0.49\linewidth}{!} {
\begin{tikzpicture} \begin{axis}[
title={}, xlabel={Number of nodes $n$}, ylabel={Wallclock time ($s$)}, grid=major,
legend entries={Global update decoding},
xtick={1280, 5120, 10240, 20480},
xticklabels={$1280$, $5120$, $10240$, $20480$},
xtick scale label code/.code={},
ytick={1.11,5.63,13.93,154.45},
yticklabels={{\small 1.11 }, {\scriptsize 5.63 \qquad \qquad }, {\small 13.93},154.45},
legend style={at={(0.02,0.98)},anchor=north west}
]
\addplot[red,mark=star] table [x=n, y=t-gc-global, col sep=comma] {./Data/Runtime/Runtime-gc-global.csv};
\end{axis}
\end{tikzpicture}
}

\caption{Runtime scaling of graph coloring inference on ER graphs for local and global decoding.}\label{fig:gc-runtime}
\end{figure*}

\Cref{fig:gc-runtime} shows how the runtime of our graph coloring model scales with the size of the graph on S-ER graphs with the edge probability $p$ as in validation according to \Cref{eqn:er-p}. Each datapoint is the mean of $10$ samples. We can see that while our model with the local decoding has nearly linear runtime scaling, the global decoding (from \Cref{sec:global-decoding}) does not scale well to graphs with more than $10{,}000$ nodes.
In particular, it takes less than $1$ second to color ER graphs with $1{,}280$ vertices ($>5{,}000$ edges) and less than $22.1$ seconds to color graphs with $20{,}480$ vertices ($>100{,}000$ edges). 
If the local decoding is replaced with globally updating the attention weights, the runtime increases significantly to more than $154$ seconds to color the same graphs with $20{,}480$ vertices.

Similar results hold for MVC: it takes less than $1$ second to compute a vertex cover for a graph with $1{,}000$ vertices and less than $40$ seconds to compute a vertex cover for a graph with $20{,}480$ vertices ($>100{,}000$ edges). With global decoding, it takes $180.9$ to compute a vertex cover.

\section{Additional Proofs}\label{apx:proofs}

\subsection{The node labeling MDP }\label{apx:proofs:mdp}

\begin{proof}[Proof of \Cref{lem:mdp}]
Consider a sequence of actions $(v_1, \ell_1)$, $\dotsc, (v_n, \ell_n)$ ending in a terminal state. For all $t$, the prefix $(v_1, \ell_1)$, $\dotsc, (v_t, \ell_t)$ of this sequence corresponds to a partial node labeling $c'$ (by viewing the sequence of node-label pairs as describing a function from nodes to labels). By construction of the MDP, labeling node $v_{i+1}$ with $\ell_{t+1}$ passes the extensibility test for $c'$. Hence the node labelling $c$ represented by $(v_1, \ell_1), \dotsc, (v_n, \ell_n)$ is feasible. By constriction, the return of the episode is $-f(c)$, where $f(c)$ is the cost of node labeling $c$.

Conversely, consider a feasible solution $c$ with cost $f(c)$. Then, by definition of feasibility (\Cref{sec:experiments:ablation}), there is a sequence $(v_1, \ell_1)$, $\dotsc, (v_n, \ell_n)$ of node-label pairs such that for all $t\geq 0$ the partial node labeling given by $(v_1, \ell_1), \dotsc, (v_t, \ell_t)$ passes the extensibility test for node $v_{t+1}$ and label $\ell_{t+1}$. Hence, the sequence of node-label pairs is also a sequence of actions in the MDP leading to a terminal state. The return for this episode is $-f(c)$.
\end{proof}

\subsection{Optimality of the labeling rule}\label{apx:proofs:label}

\begin{proof}[Proof of \Cref{lem:color-order}]
Let $G$ be some graph with chromatic number $\chi(G) = k$ and $c^{*}$ be a mapping that colors $G$ optimally. 

We partition $V$ into color classes \( C_i = \{v \,|\, c^{*}(v) = i\} \) such that all nodes with color $i$ are in $C_i$. Now, we build an ordering by consecutively taking all nodes from $C_1$, then all nodes from $C_2$ and so on. Choosing the smallest color that passes the extensibility test will produce an optimal coloring for such an order of nodes. This coloring might be different from the one of $c^{*}$. This is, because a node in $C_i$ might have no conflicts with some color $j < i$ and therefore this node will be assigned color $j$. However, by assumption all nodes that have color $j$ in $c^*$ are already colored and a node from $C_i$ can have at most $i-1$ conflicting colors in its neighborhood. Hence, the algorithm will produce a coloring of at most $k$ colors, which we assumed to be optimal.
\end{proof}

\begin{proof}[Proof of \Cref{lem:mvc-order}]
Let $S$ be the set over nodes with label $1$ in a minimum vertex cover of $G$. Order these nodes first (in an arbitrary relative order), then order the remaining nodes in $V-S$ after these nodes (in an arbitrary relative order). Labeling the nodes in this order produces a minimum vertex cover of $G$.
\end{proof}

\section{List of combinatorial node labeling problems}\label{apx:node-labeling}

We provide an extensive list of classic graph optimization problems framed as node labeling problems. Note that there can be multiple equivalent formulations. For some problems, we consider a \emph{weighed graph} $G$ with weight function $w:E\mapsto \R^{+}$, we write $w(u,v)$ the weight of an edge $\{u,v\}$. For a set of nodes $S$, we denote the subgraph of $G$ induced by $S$ with $G[S]$.

The problems in \Cref{tab:node-label-variable} require a partition of the nodes as their solution. These can be represented as node labeling problems by giving each partition its unique label. For many of the problems, the number of used labels determines the cost function.

The problems in \Cref{tab:node-label-permute} require a path (or a sequences of nodes) as their solution, which we represent as node labeling problems by having the label indicate the position in the path (or sequence).

The problems in \Cref{tab:node-label-binary} require a set of nodes as their solution. These can be represented as node labeling problems by giving the nodes in the solution set the label $1$ and the nodes not in the solution set the label $0$. The cost function is closely related to the number of nodes with label $1$ for most of these problems.

\begin{table*}
\caption{Node labeling problems which partition the nodes into $2$ or more sets.}\label{tab:node-label-variable}
\renewcommand{\arraystretch}{1.5}
\begin{tabular}{lll}
\toprule Problem & Extensibility Test $T(V'\times\mathcal{L}, v, l)$  & Cost function $f$ \\ \midrule
Balanced $k$-partition~\cite{DBLP:journals/bstj/KernighanL70} & \makecell[tl]{There are no more than $\lceil \frac{n}{k} \rceil$\\nodes with the same label\\and at most $k$ labels.} & $ \sum_{\{u, v\} \in E, l(u) \neq l(v)}  w(u, v)$   \\
\makecell[tl]{Balanced $(k, 1+\epsilon)$-partition  \cite{DBLP:journals/bstj/KernighanL70}} &\makecell[tl]{There are no more than $\lceil \frac{n(1+\epsilon)}{k} \rceil $\\nodes with the same label\\and at most $k$ labels.}  & $ \sum_{\{u, v\} \in E, l(u) \neq l(v)}  w(u, v)$   \\
\makecell[tl]{Minimum $k$-cut \cite{DBLP:journals/jacm/KargerS96}} &  \makecell[tl]{ $k-|V|-|V'|-1 \leq |\mathcal{L}\cup\{v\}|$\\and $|\mathcal{L}\cup\{v\}|\leq k$}  & $ \sum_{\{u, v\} \in E, l(u) \neq l(v)}  w(u, v)$  \\
Clique cover~\cite{Kar72} & Every label induces a clique & Number of labels \\
Domatic number~\cite{DBLP:journals/dm/HedetniemiL90a} & \makecell[tl]{Every label induces a\\dominating set of $G[V' \cup \{v\}]$} & Negative number of labels \\
Graph coloring~\cite{jensen1995graph,Kar72} & \makecell[tl]{No neighbor of $v$ has label $l$} & Number of labels \\
Graph co-coloring~\cite{jensen1995graph} & \makecell[tl]{The nodes with label $l$ induce\\ an independent set in $G$ \\ or the complement of $G$} & Number of labels \\
$k$-defective coloring~\cite{DBLP:journals/jgt/CowenCW86} & \makecell[tl]{No node has more than $k$\\neighbors with label $l$} & Number of labels \\ \bottomrule
\end{tabular}
\end{table*}

\begin{table*}
\renewcommand{\arraystretch}{1.7}
\caption{Node labeling problems where the labels encode a sequence of nodes.}\label{tab:node-label-permute}
\begin{tabular}{lll}
\toprule Problem & Extensibility Test $T(V'\times\mathcal{L}, v, l)$  & Cost function $f$ \\ \midrule
\makecell[tl]{Traveling salesman problem \cite{DBLP:journals/ior/DantzigFJ54}} &  \makecell[tl]{$l=\max(\mathcal{L})+1$ and \\ $v$ is a neighbor of the node in $\mathcal{L}$\\ with label $\max(\mathcal{L})$} & $\sum_{(u, v)\in E, l(v)=l(u)+1} w(u,v)$\\
\makecell[tl]{Tree decomposition \cite{DBLP:conf/sofsem/Bodlaender05}} & $l=\max(\mathcal{L})+1$ & \makecell[tl]{For a node $v_i$ with label $i$, add edges\\to $G$ until $v_i$ forms a clique with\\its higher-labelled neighbors.\\The cost is the largest number \\  of higher-labelled neighbors\\ in the augmented graph~\cite{DBLP:conf/sofsem/Bodlaender05}.} \\
Longest path~\cite{DBLP:journals/algorithmica/KargerMR97} & $l=\max(\mathcal{L})+1$ & \makecell[tl]{Maximum number of nodes with \\consecutive labels that induce a path}\\
\bottomrule
\end{tabular}
\end{table*}

\begin{table*}
\caption{Node labeling problems with binary labels. \emph{For all these problems}, the extensibility test passes only if the label is $0$ or $1$ (and the additional requirements listed below are satisfied).}\label{tab:node-label-binary}
\renewcommand{\arraystretch}{1.7}
\begin{tabular}{lll}
\toprule Problem & Extensibility Test $T(V'\times\mathcal{L}, v, l)$  & Cost function $f$ \\ \midrule
Maximum cut~\cite{Kar72} & At least one node has label $1$ &  $-|\{ \{u, v\} \in E, l(u) \neq l(v) \}|$ \vspace{0.5em}\\

Sparsest cut~\cite{DBLP:journals/jacm/AroraRV09}  &  At least one node has label $1$ & $\frac{|\{ \{u, v\} \in E, \ l(u) \neq l(v) \}|}{|\{ v \in V, \ l(v)=1\}|}$  \vspace{0.5em} \\

\makecell[tl]{Maximum independent set \cite{DBLP:journals/siamcomp/TarjanT77,DBLP:journals/jal/Robson86}} & \makecell[tl]{The subgraph induced by\\the nodes with label $1$ is an\\independent set} & $-|\{ v \in V, l(v)=1\}|$ \\
\makecell[tl]{Minimum vertex cover (node cover)~\cite{Kar72}} &  \makecell[tl]{The subgraph induced by\\the nodes with label $1$ is a\\vertex cover of $G[V'\cup \{v\}]$} & $|\{ v \in V, l(v)=1\}|$ \\
Maximum clique~\cite{DBLP:conf/dmtcs/TomitaS03} & \makecell[tl]{The subgraph induced by\\the nodes with label $1$ is a\\clique}  & $-|\{ v \in V, l(v)=1\}|$ \\
\makecell[tl]{Minimum feedback node set \cite{Kar72}} &  \makecell[tl]{$G[\{u\in V'\cup \{v\}, l(u)=0\}]$ \\is a forest}  & $|\{ v \in V, l(v)=1\}|$\\
Metric dimension~\cite{harary1976metric} & \makecell[tl]{The nodes in $V'\cup\{v\}$ are\\uniquely identified by their\\distances to nodes\\with label $1$}  & $|\{ v \in V, l(v)=1\}|$\\
\makecell[tl]{Minimum dominating set \cite{DBLP:journals/dm/HedetniemiL90a}} & \makecell[tl]{The nodes with label $1$ form a\\dominating set of $G[V'\cup \{v\}]$}  & $|\{ v \in V, l(v)=1\}|$\\
\makecell[tl]{Minimum connected dominating set \cite{DBLP:journals/dm/HedetniemiL90a}} & \makecell[tl]{The nodes with label $1$ form a\\connected dominating\\ set of $G[V'\cup \{v\}]$}  & $|\{ v \in V, l(v)=1\}|$\\
\bottomrule
\end{tabular}
\end{table*}

\end{document}